\documentclass[10pt,journal,compsoc]{IEEEtran}

\usepackage{iftex}
\ifXeTeX
  \usepackage{fontspec}
\fi
\usepackage{amsmath,amssymb}
\usepackage{graphicx}
\graphicspath{{figures/}}
\usepackage{booktabs}
\usepackage[nocompress]{cite}
\usepackage{multirow}
\usepackage{array}
\usepackage{ragged2e}
\usepackage[nospread,noshrink]{flushend}
\usepackage{placeins}
\usepackage{url}
\usepackage{enumitem}
\usepackage{xcolor}
\definecolor{tpamiLinkRed}{RGB}{234,51,33}
\definecolor{tpamiLinkBlue}{RGB}{0,74,155}
\definecolor{tpamiCiteGreen}{RGB}{45,190,85}
\definecolor{tpamiUrlMagenta}{RGB}{210,0,115}
\usepackage[
  colorlinks=true,
  linkcolor=tpamiLinkRed,
  citecolor=tpamiCiteGreen,
  urlcolor=tpamiUrlMagenta
]{hyperref}
\usepackage{xstring}
\usepackage{fontawesome5}
\usepackage{orcidlink}
\usepackage{colortbl}
\usepackage{pifont}
\usepackage{tikz}
\usetikzlibrary{arrows.meta,calc,positioning}
\usepackage[edges]{forest}


\title{\LARGE OCR in the Foundation Model Era:\\
From Specialist Models to Unified Models}

\author{Zhiheng Li\,\orcidlink{0009-0004-4020-0637}, Zongyang Ma\,\orcidlink{0009-0003-1980-9379}, Jiaxian Chen\,\orcidlink{0009-0003-4552-8378}, Cheng Cui\,\orcidlink{0009-0007-9008-922X}, Yi Liu\,\orcidlink{0000-0001-7575-2299}, Ziqi Zhang\,\orcidlink{0000-0002-5937-183X}, Chunfeng Yuan\,\orcidlink{0000-0003-2219-4961}, Bing Li\,\orcidlink{0000-0002-5888-6735}, Xiang Bai\,\orcidlink{0000-0002-3449-5940}, and Weiming Hu\,\orcidlink{0000-0001-9237-8825}%
\IEEEcompsocitemizethanks{
\IEEEcompsocthanksitem Z. Li, Z. Ma, Z. Zhang, C. Yuan, B. Li, and W. Hu are with the State Key Laboratory of Multimodal Artificial Intelligence Systems, Institute of Automation, Chinese Academy of Sciences, Beijing 100190, China.
\IEEEcompsocthanksitem Z. Li, C. Yuan, and B. Li are also with the School of Artificial Intelligence, University of Chinese Academy of Sciences, Beijing 100049, China.
\IEEEcompsocthanksitem Z. Li, C. Yuan, B. Li, and W. Hu are also with the Beijing Key Laboratory of Super Intelligent Security of Multi-Modal Information, Beijing, China.
\IEEEcompsocthanksitem J. Chen is with Northwestern Polytechnical University, Xi'an 710072, China.
\IEEEcompsocthanksitem C. Cui and Y. Liu are with Baidu Inc., Beijing, China.
\IEEEcompsocthanksitem X. Bai is with the School of Software Engineering, Huazhong University of Science and Technology, Wuhan, China.
\IEEEcompsocthanksitem Corresponding author: Chunfeng Yuan.
}
}

\hypersetup{
  pdftitle={OCR in the Foundation Model Era: From Specialist Models to Unified Models},
  pdfauthor={Zhiheng Li, Zongyang Ma, Jiaxian Chen, Cheng Cui, Yi Liu, Ziqi Zhang, Chunfeng Yuan, Bing Li, Xiang Bai, Weiming Hu}
}

\begin{document}
\raggedbottom

\IEEEtitleabstractindextext{%
\begin{abstract}
\justifying
Optical Character Recognition (OCR) is among the oldest and most widely deployed applications of artificial intelligence. This survey reviews its development, focusing on the Foundation Model Era. We define OCR through \emph{faithful transcription}: within a specified visual scope, OCR should completely reconstruct the text, notation, and document structure observable in the source image, and every textual, notational, or structural element in the output must be verifiable from visual evidence in that image. We trace the development from specialist OCR models toward unified models, examining joint text detection and recognition, structured sequence generation, and the use of pretrained vision--language models (VLMs). We organize contemporary OCR systems into five coexisting families: modular toolkits and task-specific models, specialized document VLMs, hybrid workflows, general-purpose VLMs, and commercial products and managed services. We compare their architectures, data, and training, examining hallucination and omission, insufficient spatial grounding, and inference cost in generative OCR. We review evaluation from character and scene text recognition to parsing individual structural elements and full-page transcription, highlighting gaps in consistent output definitions, attribution of performance to individual components, and coverage of real capture conditions and visible content and structure. Finally, we summarize the challenges facing OCR and discuss directions for future research. We continuously track related work at \url{https://github.com/zhihengli-casia/Awesome-OCR-in-the-Foundation-Model-Era}.
\end{abstract}

}

\maketitle
\IEEEdisplaynontitleabstractindextext
\IEEEpeerreviewmaketitle

\IEEEraisesectionheading{\section{Introduction}\label{sec:intro}}


\IEEEPARstart{O}{ptical} Character Recognition (OCR) is among the oldest and most widely deployed applications of artificial intelligence. Efforts to read printed text by machine date back to the 1910s~\cite{Schantz1982_OCRHistory}, and by the 1950s dedicated OCR devices matched optically sensed character patterns against stored templates to produce machine-readable codes~\cite{Mori1992_OCRSurvey}.

Over the following decades, OCR moved well beyond its original reliance on dedicated hardware. The targets of recognition expanded along several axes: from printed to handwritten text~\cite{Plamondon2000_Handwriting}, from scanned pages to text captured in natural scenes as cameras became ubiquitous~\cite{Ye2015_TextSurvey}, and from plain passages to documents with rich structures such as tables, formulas, and layouts with multiple columns~\cite{Zhong2020_PubTabNet, Deng2017_Im2Markup}. Generative models, increasingly built on pretrained vision--language models (VLMs), produce JSON, Markdown, HTML, or \LaTeX{} directly from document images~\cite{Kim2022_Donut, Wei2024_GOT}.

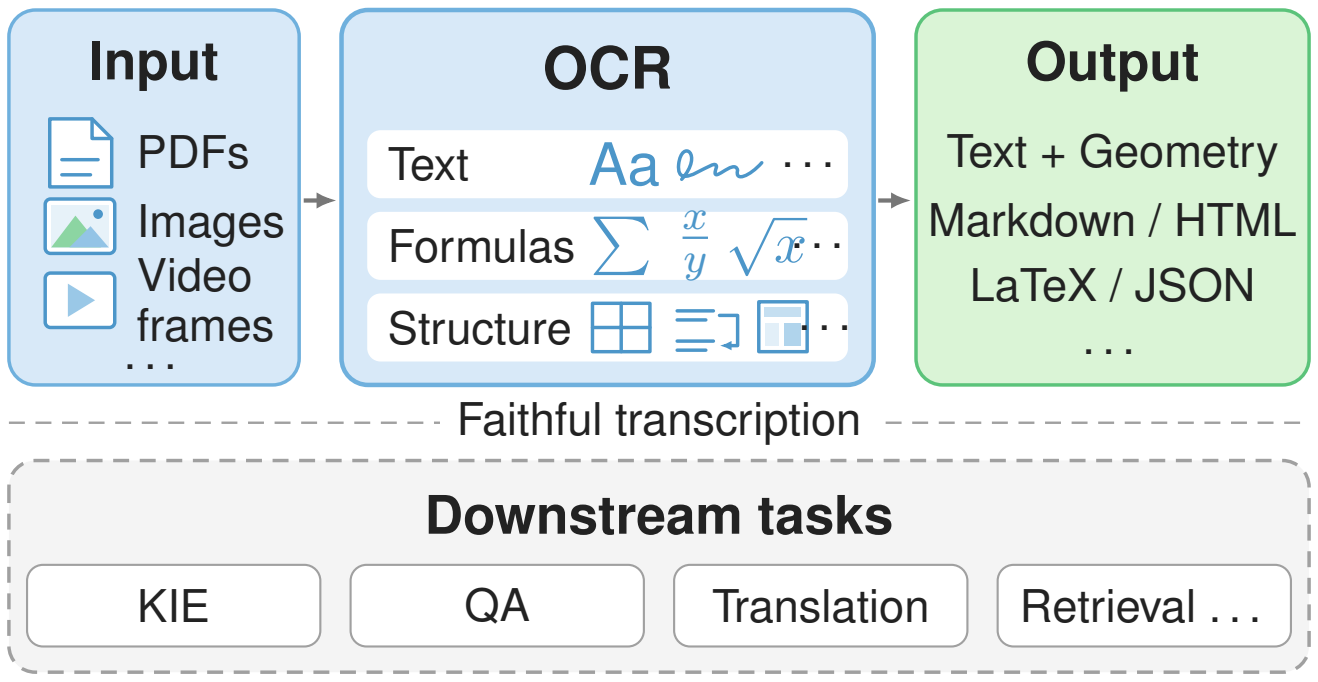
\begin{figure}[!t]
\centering
\definecolor{scopeBlue}{HTML}{6FB0DD}
\definecolor{scopeBlueDark}{HTML}{4C96C9}
\definecolor{scopeGreen}{HTML}{5BC37A}
\definecolor{scopeGray}{HTML}{777777}
\definecolor{scopeInk}{HTML}{222222}
\definecolor{scopeBlueFill}{HTML}{D8E9F8}
\definecolor{scopeGreenFill}{HTML}{DAF4D6}
\definecolor{scopeGrayFill}{HTML}{F4F4F4}
\begin{tikzpicture}[
  heading/.style={font=\sffamily\bfseries\fontsize{10pt}{11pt}\selectfont, text=scopeInk},
  labeltext/.style={font=\sffamily\fontsize{8.5pt}{9.5pt}\selectfont, text=scopeInk, align=center},
  tile/.style={labeltext, rounded corners=3pt, inner sep=2pt, minimum height=.54cm},
  flow/.style={-{Latex[length=1.35mm,width=1.1mm]},draw=scopeGray,line width=.8pt},
  sample/.style={draw=scopeBlueDark,line width=.65pt,line cap=round,line join=round}
]
\path[use as bounding box] (-4.35967,-1.99967) rectangle (4.35967,2.63);
\begin{scope}[overlay]
\draw[rounded corners=5pt,draw=scopeBlue,fill=scopeBlueFill,line width=.65pt]
  (-4.30,.02) rectangle (-2.38,2.50);
\draw[rounded corners=5pt,draw=scopeBlue,fill=scopeBlueFill,line width=.85pt]
  (-2.10,.02) rectangle (1.42,2.50);
\draw[rounded corners=5pt,draw=scopeGreen,fill=scopeGreenFill,line width=.65pt]
  (1.70,.02) rectangle (4.30,2.50);
\node[heading] at (-3.34,2.13) {Input};
\node[font=\sffamily\bfseries\fontsize{11pt}{12pt}\selectfont,text=scopeInk]
  at (-.34,2.12) {OCR};
\node[heading] at (3.00,2.13) {Output};

\draw[sample,fill=white] (-4.03,1.33) -- (-4.03,1.78) -- (-3.75,1.78)
  -- (-3.62,1.65) -- (-3.62,1.33) -- cycle;
\draw[sample] (-3.75,1.78) -- (-3.75,1.65) -- (-3.62,1.65);
\draw[sample] (-3.95,1.51) -- (-3.71,1.51);
\draw[sample] (-3.95,1.41) -- (-3.76,1.41);
\node[labeltext,anchor=west,inner sep=0pt] at (-3.46,1.56) {PDFs};
\begin{scope}[yshift=.05cm]
\draw[sample,rounded corners=1pt,fill=white] (-4.06,.84) rectangle (-3.60,1.20);
\fill[scopeBlue!17] (-4.02,.88) rectangle (-3.64,1.16);
\fill[scopeGreen!70] (-4.02,.88) -- (-3.86,1.09) -- (-3.69,.88) -- cycle;
\fill[scopeBlue!75] (-3.90,.88) -- (-3.76,1.02) -- (-3.64,.88) -- cycle;
\fill[scopeBlueDark] (-3.71,1.10) circle (.032);
\node[labeltext,anchor=west,inner sep=0pt] at (-3.46,1.02) {Images};
\end{scope}
\begin{scope}[yshift=.10cm]
\draw[sample,rounded corners=1pt,fill=white] (-4.06,.30) rectangle (-3.60,.66);
\fill[scopeBlue!70] (-3.91,.37) -- (-3.91,.59) -- (-3.73,.48) -- cycle;
\node[labeltext,align=left,anchor=west,inner sep=0pt] at (-3.46,.48) {Video\\frames};
\end{scope}
\node[labeltext,inner sep=0pt] at (-3.34,.14) {\ldots};

\foreach \y in {1.48,.94,.40}{
  \fill[white,rounded corners=2.5pt] (-1.93,\y-.225) rectangle (1.25,\y+.225);
}
\node[labeltext,anchor=west,inner sep=0pt] at (-1.80,1.48) {Text};
\node[labeltext,anchor=west,inner sep=0pt] at (-1.80,.94) {Formulas};
\node[labeltext,anchor=west,inner sep=0pt] at (-1.80,.40) {Structure};
\node[font=\sffamily\fontsize{11pt}{12pt}\selectfont,text=scopeBlueDark,inner sep=0pt]
  at (-.23,1.48) {Aa};
\draw[sample] (.15,1.38) .. controls (.30,1.55) and (.15,1.65) .. (.13,1.48)
  .. controls (.11,1.27) and (.39,1.55) .. (.38,1.46)
  .. controls (.33,1.32) and (.50,1.53) .. (.52,1.44)
  .. controls (.50,1.36) and (.60,1.40) .. (.68,1.49);
\node[labeltext,inner sep=0pt] at (1.01,1.48) {\ldots};
\node[font=\fontsize{11pt}{12pt}\selectfont,text=scopeBlueDark,inner sep=0pt]
  at (-.25,.94) {$\sum$};
\node[font=\fontsize{11pt}{12pt}\selectfont,text=scopeBlueDark,inner sep=0pt]
  at (.24,.94) {$\frac{x}{y}$};
\node[font=\fontsize{11pt}{12pt}\selectfont,text=scopeBlueDark,inner sep=0pt]
  at (.71,.94) {$\sqrt{x}$};
\node[labeltext,inner sep=0pt] at (1.07,.94) {\ldots};
\draw[sample] (-.44,.24) rectangle (-.06,.56);
\draw[sample] (-.44,.40) -- (-.06,.40) (-.25,.24) -- (-.25,.56);
\draw[sample] (.12,.51) -- (.36,.51) (.12,.45) -- (.31,.45);
\draw[sample] (.12,.31) -- (.36,.31) (.12,.25) -- (.31,.25);
\draw[sample,-{Latex[length=.8mm,width=.65mm]}]
  (.40,.48) -- (.52,.48) -- (.52,.28) -- (.40,.28);
\draw[sample] (.66,.24) rectangle (.98,.57);
\fill[scopeBlue!40] (.70,.47) rectangle (.94,.53);
\fill[scopeBlue!20] (.70,.28) rectangle (.79,.43);
\fill[scopeBlue!55] (.83,.28) rectangle (.94,.43);
\node[labeltext,inner sep=0pt] at (1.12,.40) {\ldots};

\node[labeltext] at (3.00,1.54) {Text + Geometry};
\node[labeltext] at (3.00,1.11) {Markdown / HTML};
\node[labeltext] at (3.00,.68) {LaTeX / JSON};
\node[labeltext,inner sep=0pt] at (3.00,.24) {\ldots};
\draw[flow] (-2.35,1.24) -- (-2.13,1.24);
\draw[flow] (1.45,1.24) -- (1.67,1.24);
\draw[scopeGray!70,densely dashed,line width=.45pt] (-4.30,-.23) -- (4.30,-.23);
\node[labeltext,text=scopeInk,fill=white,inner xsep=3pt,inner ysep=0pt]
  at (0,-.23) {Faithful transcription};

\draw[rounded corners=5pt,draw=scopeGray!70,densely dashed,fill=scopeGrayFill,line width=.65pt]
  (-4.30,-.48) rectangle (4.30,-1.88);
\node[heading,text=scopeInk] at (0,-.84) {Downstream tasks};
\node[tile,fill=white,draw=scopeGray!70,minimum width=1.94cm] at (-3.21,-1.44) {KIE};
\node[tile,fill=white,draw=scopeGray!70,minimum width=1.94cm] at (-1.07,-1.44) {QA};
\node[tile,fill=white,draw=scopeGray!70,minimum width=1.94cm] at (1.07,-1.44) {Translation};
\node[tile,fill=white,draw=scopeGray!70,minimum width=1.94cm] at (3.21,-1.44) {Retrieval \ldots};
\end{scope}
\end{tikzpicture}
\caption{\textbf{Scope of this survey.} Within a specified visual scope, OCR aims to completely recover visible text, notation, and document structure, with every output element verifiable against the source image. Main coverage spans printed, handwritten, and scene text; formulas as notation; and tables, layout, and reading order as structure. Key information extraction, question answering, translation, and retrieval are downstream tasks.}
\label{fig:ocr_scope}
\vspace{-7pt}
\end{figure}


\begin{figure*}[!t]
\centering
\includegraphics[width=\textwidth]{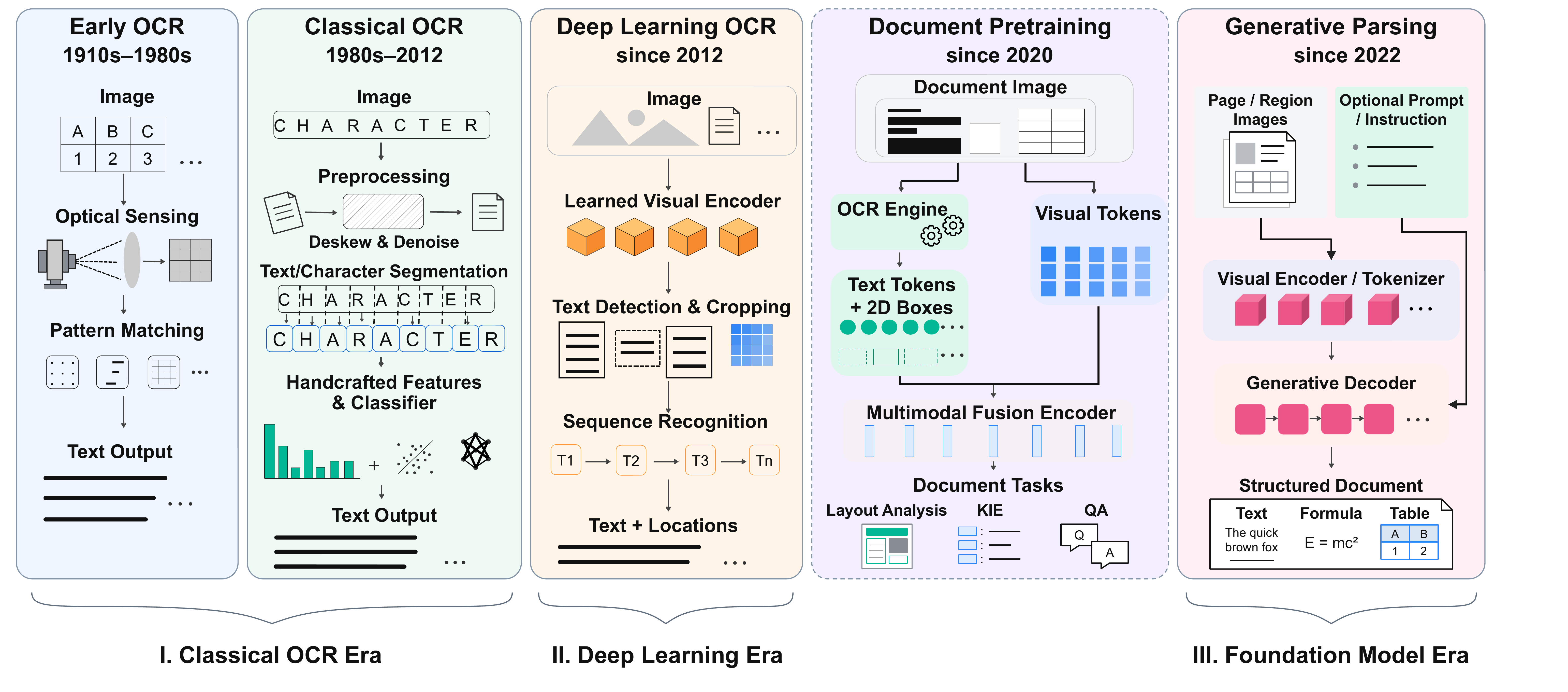}
\caption{\textbf{OCR and related document modeling across five historical tracks and three eras.} Columns show representative pipelines; braces mark three eras: \textbf{I.~Classical OCR Era}, \textbf{II.~Deep Learning Era}, and \textbf{III.~Foundation Model Era}. The dashed \textbf{Document Pretraining} box provides related background: these models combine upstream OCR text, layout, and visual features for document understanding. Generative parsing reconstructs text, formulas, and tables within a unified structured document, using formats such as Markdown, HTML, \LaTeX{}, or JSON. Dates indicate approximate emergence or historical prominence; the tracks overlap in time and several remain active in contemporary systems.}
\label{fig:ocr_evolution}
\end{figure*}

\noindent\textbf{\textit{What counts as OCR?}}
The literature uses OCR with different boundaries: technology for scanned documents~\cite{Ye2015_TextSurvey}, camera-based OCR~\cite{Chen2021_TextWild}, artificial optical signals~\cite{Wei2024_GOT}, text and visual structure recovery in holistic OCR~\cite{OCRVerse2026}, and multimodal evaluations that include information extraction and document question answering~\cite{Liu2024_OCRBench}. In this survey, we define OCR through \emph{faithful transcription}. Within a specified visual scope, OCR aims to completely reconstruct the text, notation, and document structure observable in the source image. Every textual, notational, or structural element in the output must be verifiable from visual evidence in that image (Fig.~\ref{fig:ocr_scope}). Normalization must follow a declared output convention. Our main coverage includes printed, handwritten, and scene text, mathematical formulas, tables, layout, and reading order. Chart data, flowchart connections, musical notation, and chemical structures can also be recovered from visual evidence; we discuss these as related extensions with selective coverage.

Recovering a table's cells, rows, and merged cells, or the values a chart encodes through marks and axes, reconstructs the source document; computing a statistic from those values or explaining a trend goes beyond it. Document understanding assigns semantic fields~\cite{Hwang2021_SPADE}, answers questions~\cite{Mathew2021_DocVQA}, or draws conclusions about a document~\cite{Lewis2020_RAG,Hong2024_CogAgent}. These downstream tasks have influenced OCR architectures, training, and evaluation.


\noindent\textbf{\textit{From specialist models to unified models:}}
Fig.~\ref{fig:ocr_evolution} traces three overlapping eras: the Classical OCR, Deep Learning, and Foundation Model eras. Classical methods relied on templates, handcrafted features, and statistical models. Deep learning brought learned features and sequence prediction, while pipelines continued to assemble outputs from specialist components~\cite{Du2020_PPOCR}. Joint modeling also developed: Mask TextSpotter~\cite{Lyu2018_MaskTextSpotter} integrated detection and recognition in one network, SPTS~\cite{Peng2022_SPTS} represented locations and text in one output sequence, and OmniParser~\cite{OmniParser2024} supported text spotting, table recognition, and key information extraction within a common framework.

The Foundation Model Era expanded task sharing through large-scale pretraining and structured generation. Donut~\cite{Kim2022_Donut} generated structured results for document tasks directly from images, and Nougat~\cite{Blecher2023_Nougat} adapted this approach to full-page scientific transcription. GOT-OCR~\cite{Wei2024_GOT} and HunyuanOCR~\cite{HunyuanOCR2025} used pretrained large language models (LLMs) to support multiple tasks and content types in one model. The development from specialist to unified models thus spans these eras, with increasing sharing of networks, representations, and decoders.

In practice, unified models can generate complete pages or work with layout analysis to generate regions before document assembly. Specialist models, whole-page generation, and regional generation continue to coexist; Section~\ref{sec:vlm} compares their architectures.



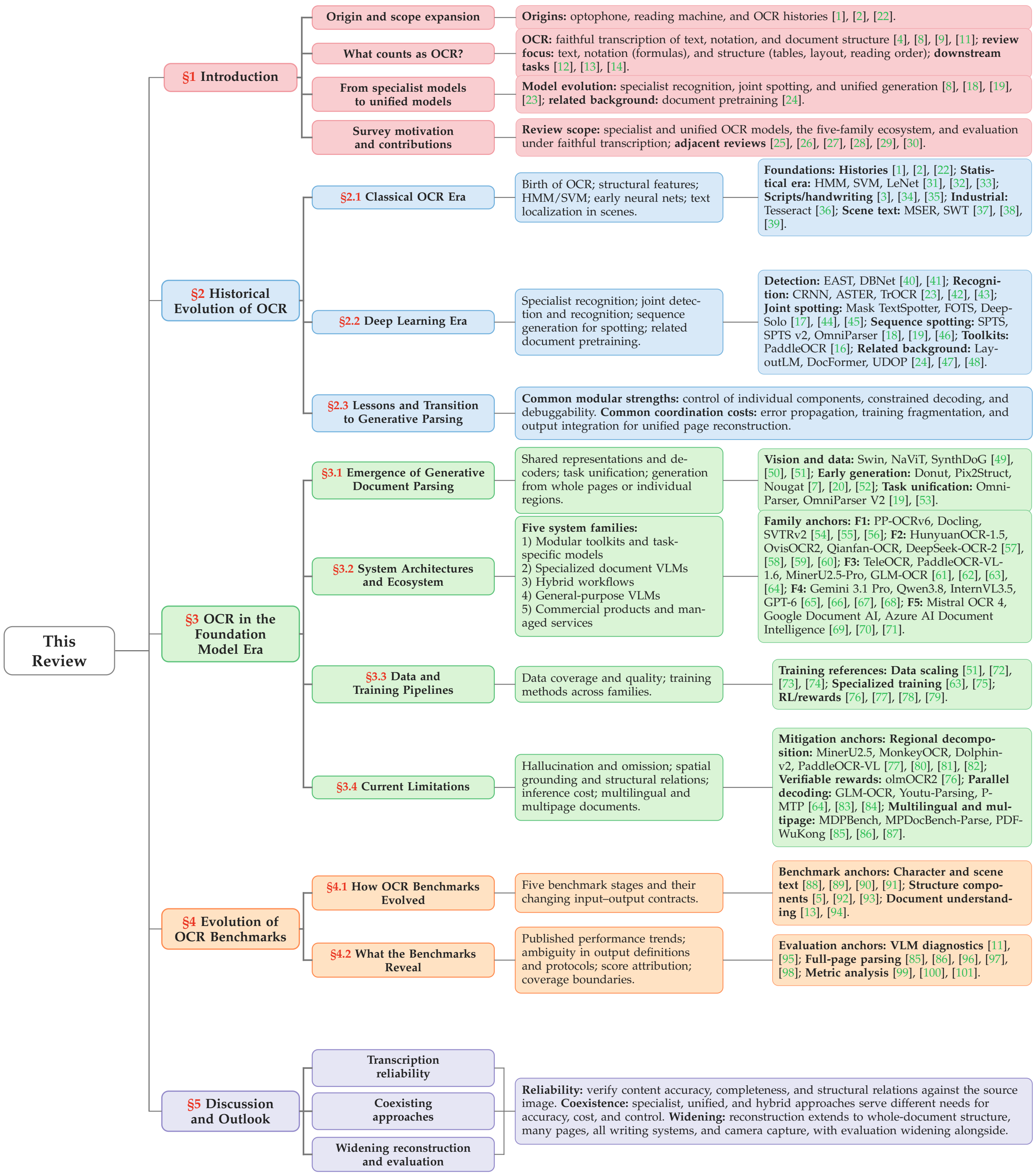
\begin{figure*}[!t]
\centering
\definecolor{introC}{HTML}{F1989F}
\definecolor{histC}{HTML}{6FB0DD}
\definecolor{vlmC}{HTML}{5BC37A}
\definecolor{expC}{HTML}{FF8C51}
\definecolor{futureC}{HTML}{938ACA}
\definecolor{concC}{HTML}{777777}
\definecolor{inkC}{HTML}{222222}
\definecolor{lineC}{HTML}{777777}
\definecolor{introCFill}{HTML}{F9CDD0}
\definecolor{histCFill}{HTML}{D8E9F8}
\definecolor{vlmCFill}{HTML}{DAF4D6}
\definecolor{expCFill}{HTML}{FFE2C3}
\definecolor{futureCFill}{HTML}{E6E6F4}
\begingroup
\newcommand{\figseclink}[3]{\hyperref[#1]{#3}}
\vspace*{-1mm}
\resizebox{0.96\textwidth}{!}{%
\begin{tikzpicture}[
  x=1cm,y=1cm,
  font=\normalfont\rmfamily,
  root/.style={rounded corners=5pt, draw=lineC!90, fill=white, line width=.65pt,
    align=center, font=\normalfont\rmfamily\bfseries\normalsize, text=inkC, inner sep=5pt},
  sec/.style={rounded corners=4pt, draw=#1, fill=#1Fill, line width=.72pt,
    align=center, font=\normalfont\rmfamily\bfseries\footnotesize, text=inkC, inner sep=4.5pt},
  sub/.style={rounded corners=3.5pt, draw=#1, fill=#1Fill, line width=.58pt,
    align=center, font=\normalfont\rmfamily\bfseries\scriptsize, text=inkC, inner sep=3.5pt},
  topic/.style={rounded corners=3.5pt, draw=#1, fill=#1Fill, line width=.48pt,
    align=left, font=\normalfont\rmfamily\scriptsize, text=inkC, inner sep=4pt},
  anchorbox/.style={rounded corners=3.5pt, draw=#1!85, fill=#1Fill, line width=.36pt,
    align=left, font=\normalfont\rmfamily\scriptsize, text=inkC, inner sep=3.6pt},
  trunk/.style={draw=lineC, line width=.68pt},
  branch/.style={draw=lineC, line width=.55pt, line cap=rect, line join=miter},
  groupbranch/.style={draw=lineC, line width=.95pt, line cap=rect, line join=miter},
]

\path[use as bounding box] (0,0.8) rectangle (22.05,26.0);

\node[root, text width=1.95cm] (root) at (1.28,12.40) {This\\Review};
\draw[trunk] (root.east) -- (3.15,12.40);
\draw[trunk] (3.15,2.80) -- (3.15,24.35);

\node[sec=introC, text width=2.45cm] (s1) at (4.85,24.35)
  {\figseclink{sec:intro}{introC}{\S1} Introduction};
\draw[branch] (3.15,24.35) -- (s1.west);
\node[sub=introC, text width=3.55cm] (s11) at (8.45,25.60)
  {Origin and scope expansion};
\node[sub=introC, text width=3.55cm] (s12) at (8.45,24.85)
  {What counts as OCR?};
\node[sub=introC, text width=3.55cm] (s13) at (8.45,24.00)
  {From specialist models\\to unified models};
\node[sub=introC, text width=3.55cm] (s14) at (8.45,23.10)
  {Survey motivation\\and contributions};
\coordinate (b1x) at ($(s1.east)!0.30!(s11.west)$);
\coordinate (b1) at (b1x |- s1.east);
\draw[groupbranch] (s1.east) -- (b1);
\draw[groupbranch] (b1 |- s11.west) -- (b1 |- s14.west);
\foreach \n in {s11,s12,s13,s14}{\draw[groupbranch] (b1 |- \n.west) -- (\n.west);}
\node[anchorbox=introC, text width=10.50cm] (a11) at (16.175,25.60)
  {\textbf{Origins:} optophone, reading machine, and OCR histories~\cite{Schantz1982_OCRHistory,Mori1992_OCRSurvey,Impedovo1991_OCRSurvey}.};
\node[anchorbox=introC, text width=10.50cm] (a12) at (16.175,24.85)
  {\textbf{OCR:} faithful transcription of text, notation, and document structure~\cite{Ye2015_TextSurvey,Chen2021_TextWild,Wei2024_GOT,Liu2024_OCRBench}; \textbf{review focus:} text, notation (formulas), and structure (tables, layout, reading order); \textbf{downstream tasks}~\cite{Mathew2021_DocVQA,Hwang2021_SPADE,Lewis2020_RAG}.};
\node[anchorbox=introC, text width=10.50cm] (a13) at (16.175,24.00)
  {\textbf{Model evolution:} specialist recognition, joint spotting, and unified generation~\cite{Shi2017_CRNN,Peng2022_SPTS,OmniParser2024,Wei2024_GOT}; \textbf{related background:} document pretraining~\cite{Xu2020_LayoutLM}.};
\node[anchorbox=introC, text width=10.50cm] (a14) at (16.175,23.10)
  {\textbf{Review scope:} specialist and unified OCR models, the five-family ecosystem, and evaluation under faithful transcription; \textbf{adjacent reviews}~\cite{Zhu2016_SceneTextSurvey,Long2021_SceneText,Liu2023_DocumentAnalysisReview,Li2025_LargeModelOCRSurvey,Shu2025_VisualTextProcessing,Fu2025_MLLMText}.};
\foreach \s/\a in {s11/a11,s12/a12,s13/a13,s14/a14}{\draw[branch] (\s.east) -- (\a.west);}

\node[sec=histC, text width=2.55cm] (s2) at (4.85,19.70)
  {\figseclink{sec:pre_vlm}{histC}{\S2} Historical\\Evolution of OCR};
\draw[branch] (3.15,19.70) -- (s2.west);
\node[sub=histC, text width=3.55cm] (s21) at (8.45,21.85)
  {\figseclink{sec:pre_vlm_classical}{histC}{\S2.1} Classical OCR Era};
\node[sub=histC, text width=3.55cm] (s22) at (8.45,19.25)
  {\figseclink{sec:pre_vlm_dl}{histC}{\S2.2} Deep Learning Era};
\node[sub=histC, text width=3.55cm] (s23) at (8.45,17.35)
  {\figseclink{sec:pre_vlm_ceiling}{histC}{\S2.3} Lessons and Transition\\to Generative Parsing};
\coordinate (b2x) at ($(s2.east)!0.30!(s22.west)$);
\coordinate (b2) at (b2x |- s2.east);
\draw[groupbranch] (s2.east) -- (b2);
\draw[groupbranch] (b2 |- s21.west) -- (b2 |- s23.west);
\foreach \n in {s21,s22,s23}{\draw[groupbranch] (b2 |- \n.west) -- (\n.west);}

\node[topic=histC, text width=4.05cm] (t21) at (12.95,21.85)
  {Birth of OCR; structural features; HMM/SVM; early neural nets; text localization in scenes.};
\node[anchorbox=histC, text width=5.45cm] (a21) at (18.70,21.85)
  {\textbf{Foundations:}
   \textbf{Histories}~\cite{Schantz1982_OCRHistory,Mori1992_OCRSurvey,Impedovo1991_OCRSurvey};
   \textbf{Statistical era:} HMM, SVM, LeNet~\cite{Rabiner1989_HMM,Cortes1995_SVM,LeCun1998_LeNet};
   \textbf{Scripts/handwriting}~\cite{Plamondon2000_Handwriting,Liu2011_CASIAHWDB,Pal2004_IndianScript};
   \textbf{Industrial:} Tesseract~\cite{Smith2007_Tesseract};
   \textbf{Scene text:} MSER, SWT~\cite{Matas2004_MSER,Epshtein2010_SWT,Neumann2012_RealTimeText}.};
\draw[branch] (s21.east) -- (t21.west);
\draw[branch] (t21.east) -- (a21.west);

\node[topic=histC, text width=4.05cm] (t22) at (12.95,19.25)
  {Specialist recognition; joint detection and recognition; sequence generation for spotting; related document pretraining.};
\node[anchorbox=histC, text width=5.45cm] (a22) at (18.70,19.25)
  {\textbf{Detection:}
   EAST, DBNet~\cite{Zhou2017_EAST,Liao2020_DBNet};
   \textbf{Recognition:}
   CRNN, ASTER, TrOCR~\cite{Shi2017_CRNN,Shi2019_ASTER,Li2023_TrOCR};
   \textbf{Joint spotting:}
   Mask TextSpotter, FOTS, DeepSolo~\cite{Lyu2018_MaskTextSpotter,Zhou2018_FOTS,Ye2023_DeepSolo};
   \textbf{Sequence spotting:}
   SPTS, SPTS v2, OmniParser~\cite{Peng2022_SPTS,Liu2023_SPTSv2,OmniParser2024};
   \textbf{Toolkits:}
   PaddleOCR~\cite{Du2020_PPOCR};
   \textbf{Related background:}
   LayoutLM, DocFormer, UDOP~\cite{Xu2020_LayoutLM,Appalaraju2021_DocFormer,Tang2023_UDOP}.};
\draw[branch] (s22.east) -- (t22.west);
\draw[branch] (t22.east) -- (a22.west);

\node[topic=histC, text width=10.50cm] (t23) at (16.175,17.35)
  {\textbf{Common modular strengths:} control of individual components, constrained decoding, and debuggability. \textbf{Common coordination costs:} error propagation, training fragmentation, and output integration for unified page reconstruction.};
\draw[branch] (s23.east) -- (t23.west);

\node[sec=vlmC, text width=2.55cm] (s3) at (4.85,12.75)
  {\figseclink{sec:vlm}{vlmC}{\S3} OCR in the\\Foundation Model Era};
\draw[branch] (3.15,12.75) -- (s3.west);
\node[sub=vlmC, text width=3.55cm] (s31) at (8.45,15.95)
  {\figseclink{sec:vlm_shift}{vlmC}{\S3.1} Emergence of Generative\\Document Parsing};
\node[sub=vlmC, text width=3.55cm] (s32) at (8.45,13.95)
  {\figseclink{sec:vlm_taxonomy}{vlmC}{\S3.2} System Architectures\\and Ecosystem};
\node[sub=vlmC, text width=3.55cm] (s33) at (8.45,11.70)
  {\figseclink{sec:vlm_design}{vlmC}{\S3.3} Data and\\Training Pipelines};
\node[sub=vlmC, text width=3.55cm] (s34) at (8.45,9.55)
  {\figseclink{sec:vlm_limits}{vlmC}{\S3.4} Current Limitations};
\coordinate (b3x) at ($(s3.east)!0.30!(s32.west)$);
\coordinate (b3) at (b3x |- s3.east);
\draw[groupbranch] (s3.east) -- (b3);
\draw[groupbranch] (b3 |- s31.west) -- (b3 |- s34.west);
\foreach \n in {s31,s32,s33,s34}{\draw[groupbranch] (b3 |- \n.west) -- (\n.west);}

\node[topic=vlmC, text width=4.05cm] (t31) at (12.95,15.95)
  {Shared representations and decoders; task unification; generation from whole pages or individual regions.};
\node[anchorbox=vlmC, text width=5.45cm] (a31) at (18.70,15.95)
  {\textbf{Vision and data:} Swin, NaViT, SynthDoG~\cite{Liu2021_SwinTransformer,Dehghani2024_NaViT,Yim2021_SynthDoG};
   \textbf{Early generation:} Donut, Pix2Struct, Nougat~\cite{Kim2022_Donut,Lee2023_Pix2Struct,Blecher2023_Nougat};
   \textbf{Task unification:} OmniParser, OmniParser V2~\cite{OmniParser2024,Yu2026_OmniParserV2}.};
\draw[branch] (s31.east) -- (t31.west);
\draw[branch] (t31.east) -- (a31.west);

\node[topic=vlmC, text width=4.05cm] (t32) at (12.95,13.95)
  {\textbf{Five system families:}\\
   1) Modular toolkits and task-specific models\\
   2) Specialized document VLMs\\
   3) Hybrid workflows\\
   4) General-purpose VLMs\\
   5) Commercial products and managed services};
\node[anchorbox=vlmC, text width=5.45cm] (a32) at (18.70,13.95)
  {\textbf{Family anchors:}
   \textbf{F1:} PP-OCRv6, Docling, SVTRv2~\cite{PPOCRv6_2026,Docling2024,Du2024_SVTRv2};
   \textbf{F2:} HunyuanOCR-1.5, OvisOCR2, Qianfan-OCR, DeepSeek-OCR-2~\cite{HunyuanOCR15_2026,OvisOCR2_2026,QianfanOCR2026,DeepSeekOCR2_2026};
   \textbf{F3:} TeleOCR, PaddleOCR-VL-1.6, MinerU2.5-Pro, GLM-OCR~\cite{NaviDCOCR2026,PaddleOCRVL16_2026,MinerU2025,Duan2026_GLMOCR};
   \textbf{F4:} Gemini~3.1 Pro, Qwen3.8, InternVL3.5, \mbox{GPT-6}~\cite{Gemini31Pro_2026,Qwen38_2026,InternVL35_2025,GPT6_2026};
   \textbf{F5:} Mistral OCR~4, Google Document AI, Azure AI Document Intelligence~\cite{MistralOCR4_2026,Google2025_DocumentAI,Azure2025_DocumentIntelligence}.};
\draw[branch] (s32.east) -- (t32.west);
\draw[branch] (t32.east) -- (a32.west);

\node[topic=vlmC, text width=4.05cm] (t33) at (12.95,11.70)
  {Data coverage and quality; training methods across families.};
\node[anchorbox=vlmC, text width=5.15cm] (a33) at (18.85,11.70)
  {\textbf{Training references:}
   \textbf{Data scaling}~\cite{Baek2019_WhatIsWrong,Jiang2023_MAERec,Yim2021_SynthDoG,Radford2021_CLIP};
   \textbf{Specialized training}~\cite{MinerU2025,MonkeyOCRv2_2026};
   \textbf{RL/rewards}~\cite{olmOCR2_2025,PaddleOCRVL2026,Shao2024_GRPO,FDRL2026}.};
\draw[branch] (s33.east) -- (t33.west);
\draw[branch] (t33.east) -- (a33.west);

\node[topic=vlmC, text width=4.05cm] (t34) at (12.95,9.55)
  {Hallucination and omission; spatial grounding and structural relations; inference cost; multilingual and multipage documents.};
\node[anchorbox=vlmC, text width=5.15cm] (a34) at (18.85,9.55)
  {\textbf{Mitigation anchors:}
   \textbf{Regional decomposition:} MinerU2.5, MonkeyOCR, Dolphin-v2, PaddleOCR-VL~\cite{Niu2025_MinerU25,MonkeyOCR2025,Dolphin2025,PaddleOCRVL2026};
   \textbf{Verifiable rewards:} olmOCR2~\cite{olmOCR2_2025};
   \textbf{Parallel decoding:} GLM-OCR, Youtu-Parsing, P-MTP~\cite{Duan2026_GLMOCR,YoutuParsing2026,Xiang2026_PMTP};
   \textbf{Multilingual and multipage:} MDPBench, MPDocBench-Parse, PDF-WuKong~\cite{MDPBench2026,Zhou2026_MPDocBenchParse,Xie2026_PDFWuKong}.};
\draw[branch] (s34.east) -- (t34.west);
\draw[branch] (t34.east) -- (a34.west);

\node[sec=expC, text width=2.55cm] (s4) at (4.85,6.60)
  {\figseclink{sec:experiments}{expC}{\S4} Evolution of\\OCR Benchmarks};
\draw[branch] (3.15,6.60) -- (s4.west);
\node[sub=expC, text width=3.55cm] (s41) at (8.45,7.35)
  {\figseclink{sec:exp_benchmarks}{expC}{\S4.1} How OCR Benchmarks\\Evolved};
\node[sub=expC, text width=3.55cm] (s42) at (8.45,5.95)
  {\figseclink{sec:exp_crossdomain}{expC}{\S4.2} What the Benchmarks\\Reveal};
\coordinate (b4x) at ($(s4.east)!0.30!(s42.west)$);
\coordinate (b4) at (b4x |- s4.east);
\draw[groupbranch] (s4.east) -- (b4);
\draw[groupbranch] (b4 |- s41.west) -- (b4 |- s42.west);
\foreach \n in {s41,s42}{\draw[groupbranch] (b4 |- \n.west) -- (\n.west);}
\node[topic=expC, text width=4.05cm] (t41) at (12.95,7.35)
  {Five benchmark stages and their changing input--output contracts.};
\node[anchorbox=expC, text width=5.15cm] (a41) at (18.85,7.35)
  {\textbf{Benchmark anchors:}
   \textbf{Character and scene text}~\cite{Wang2011_SVT,Karatzas2015_ICDAR,Chng2017_TotalText,Liu2019_CTW1500};
   \textbf{Structure components}~\cite{Zhong2020_PubTabNet,Zhong2019_PubLayNet,UniMERNet2024};
   \textbf{Document understanding}~\cite{Mathew2021_DocVQA,Masry2022_ChartQA}.};
\node[topic=expC, text width=4.05cm] (t42) at (12.95,5.95)
  {Published performance trends; ambiguity in output definitions and protocols; score attribution; coverage boundaries.};
\node[anchorbox=expC, text width=5.15cm] (a42) at (18.85,5.95)
  {\textbf{Evaluation anchors:}
   \textbf{VLM diagnostics}~\cite{Liu2024_OCRBench,Liu2025_OCRBenchv2};
   \textbf{Full-page parsing}~\cite{OmniDocBench2025,Real5_2026,MDPBench2026,PureDocBench2026,Zhou2026_MPDocBenchParse};
   \textbf{Metric analysis}~\cite{Cui2026_RTDocLayout,Kumar2022_TEDSIOU,Wang2025_CDM}.};
\draw[branch] (s41.east) -- (t41.west);
\draw[branch] (t41.east) -- (a41.west);
\draw[branch] (s42.east) -- (t42.west);
\draw[branch] (t42.east) -- (a42.west);

\node[sec=futureC, text width=2.55cm] (s5) at (4.85,2.80)
  {\figseclink{sec:challenges}{futureC}{\S5} Discussion\\and Outlook};
\draw[branch] (3.15,2.80) -- (s5.west);
\node[sub=futureC, text width=3.55cm] (s51) at (8.45,3.70)
  {Transcription\\reliability};
\node[sub=futureC, text width=3.55cm] (s52) at (8.45,2.80)
  {Coexisting\\approaches};
\node[sub=futureC, text width=3.55cm] (s53) at (8.45,1.90)
  {Widening reconstruction\\and evaluation};
\node[topic=futureC, text width=10.50cm] (t5) at (16.175,2.80)
  {\textbf{Reliability:} verify content accuracy, completeness, and structural relations against the source image. \textbf{Coexistence:} specialist, unified, and hybrid approaches serve different needs for accuracy, cost, and control. \textbf{Widening:} reconstruction extends to whole-document structure, many pages, all writing systems, and camera capture, with evaluation widening alongside.};
\coordinate (b5x) at ($(s5.east)!0.30!(s51.west)$);
\coordinate (b5) at (b5x |- s5.east);
\draw[groupbranch] (s5.east) -- (b5);
\draw[groupbranch] (b5 |- s51.west) -- (b5 |- s53.west);
\foreach \n in {s51,s52,s53}{\draw[groupbranch] (b5 |- \n.west) -- (\n.west);}
\draw[branch] (s51.east) -- (10.55,3.70);
\draw[branch] (s52.east) -- (10.55,2.80);
\draw[branch] (s53.east) -- (10.55,1.90);
\draw[branch] (10.55,3.70) -- (10.55,1.90);
\draw[branch] (10.55,2.80) -- (t5.west);

\end{tikzpicture}%
}
\endgroup
\vspace{-17pt}
\caption{\textbf{Review framework and evidence map.} The figure connects OCR's historical evolution and five contemporary system families with benchmark evidence, current limitations, and the three expectations of the outlook. Citations identify representative references.}
\label{fig:paper_structure}
\vspace{-12pt}
\end{figure*}


\begin{figure*}[!t]
\centering
\includegraphics[width=\textwidth]{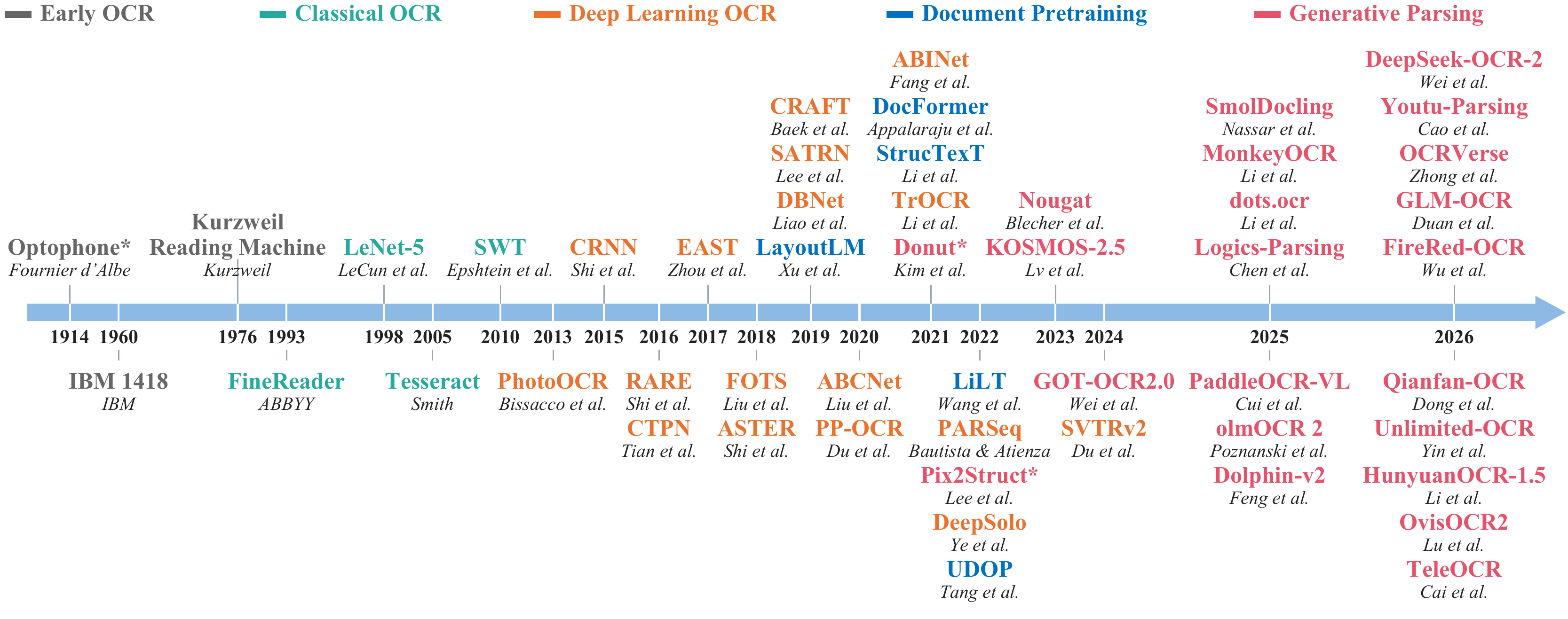}
\caption{Representative milestones in OCR and related document modeling, following the five tracks in Fig.~\ref{fig:ocr_evolution}. Years mark publications, public announcements, or system releases; horizontal spacing is schematic. Asterisks identify precursors: the optophone as a reading aid, and Donut and Pix2Struct in document understanding and image-to-sequence pretraining. Section~\ref{sec:pre_vlm} covers classical and deep learning methods; Section~\ref{sec:vlm} examines unified models and generative parsing.}
\label{fig:dl_timeline}
\nocite{Lee2020_SATRN,Li2021_StrucTexT,Wang2022_LiLT,Bautista2022_PARSeq}
\end{figure*}

\looseness=0
\noindent\textbf{\textit{Why does OCR need a new survey?}}
Mori et al.~\cite{Mori1992_OCRSurvey} reviewed the historical development of OCR. Ye and Doermann~\cite{Ye2015_TextSurvey}, Zhu et al.~\cite{Zhu2016_SceneTextSurvey}, Long et al.~\cite{Long2021_SceneText}, and Chen et al.~\cite{Chen2021_TextWild} surveyed scene text detection and recognition across different stages of its development. Liu et al.~\cite{Liu2023_DocumentAnalysisReview} provided a broader review of intelligent document analysis and recognition. Recent surveys examine OCR with large models~\cite{Li2025_LargeModelOCRSurvey}, document intelligence and parsing~\cite{Huang2024_DocAISurvey,Zhang2024_DocParsing,Li2026_DocumentMicroSalon}, text-centric image understanding~\cite{Zhang2023_TextCentricReview}, visual text processing and manipulation~\cite{Shu2024_VisualTextLowLevel,Shu2025_VisualTextProcessing}, or general vision--language models that include OCR among many capabilities~\cite{Fu2025_MLLMText}. They adopt different boundaries for OCR.

\looseness=0
In the Foundation Model Era, a single system can perform transcription, structural recovery, and downstream understanding, while specialist models, unified models, and hybrid workflows continue to coexist. This survey connects research on specialist OCR and unified models to explain how OCR has changed in the Foundation Model Era and which problems warrant further research. Fig.~\ref{fig:paper_structure} outlines the survey's structure.

\looseness=0
\noindent\textbf{\textit{Contributions:}}
(i) We systematically review the major stages and technical transitions of OCR, tracing its development from specialist models toward unified models with a focus on the Foundation Model Era.
(ii) We define OCR through faithful transcription and organize contemporary systems into five families, comparing their architectures, data, and training.
(iii) We trace OCR evaluation from text recognition to full-page parsing, analyze recognition and structural recovery using published results, and identify gaps in output definitions, score attribution, and coverage of multilingual content, long documents, and real capture conditions.
(iv) Focusing on faithful transcription, we discuss output reliability, the coexistence of specialist, unified, and hybrid approaches, and the expanding scope of content recovery.

This survey aims to help newcomers understand OCR and provide experienced researchers with a reference for recent advances and open problems.

\section{Historical Evolution of OCR}\label{sec:pre_vlm}

This section traces how OCR recovered visible text and its location, and how these predictions became increasingly integrated for document reconstruction. Detailed surveys cover these eras~\cite{Mori1992_OCRSurvey,Ye2015_TextSurvey,Zhu2016_SceneTextSurvey,Long2021_SceneText,Chen2021_TextWild,Liu2023_DocumentAnalysisReview}; Figs.~\ref{fig:ocr_evolution} and~\ref{fig:dl_timeline} summarize their evolution.

\subsection{Classical OCR Era}\label{sec:pre_vlm_classical}

Early optical reading devices included the optophone (1914), which encoded printed patterns as tones for a listener to interpret, and Tauschek's template-matching machine (1929). Commercial OCR later used standardized fonts and controlled imaging~\cite{Schantz1982_OCRHistory,Mori1992_OCRSurvey}. HMMs~\cite{Rabiner1989_HMM} modeled character sequences, SVMs~\cite{Cortes1995_SVM} classified characters, and LeNet~\cite{LeCun1998_LeNet} learned recognition features from pixels. FineReader~\cite{ABBYY2024_FineReader} and Tesseract~\cite{Smith2007_Tesseract} supported document OCR, while recognition expanded to handwriting and non-Latin scripts~\cite{Plamondon2000_Handwriting,Pal2004_IndianScript,Liu2011_CASIAHWDB}.

Scene text introduced uncontrolled typography, backgrounds, and lighting. MSER~\cite{Matas2004_MSER} located stable regions, SWT~\cite{Epshtein2010_SWT} exploited consistent stroke width, and Neumann and Matas~\cite{Neumann2012_RealTimeText} connected localization with recognition. PhotoOCR~\cite{Bissacco2013_PhotoOCR} combined handcrafted features with a deep character classifier, marking a transition toward learned representations.

\subsection{Deep Learning Era}\label{sec:pre_vlm_dl}

Deep learning replaced handcrafted features with learned representations~\cite{Krizhevsky2012_AlexNet,He2016_ResNet}. In detection, CTPN~\cite{Tian2016_CTPN} linked local text proposals, EAST~\cite{Zhou2017_EAST} regressed oriented regions, PSENet~\cite{Wang2019_PSENet} and CRAFT~\cite{Baek2019_CRAFT} supported irregular text through segmentation, and DBNet~\cite{Liao2020_DBNet} made binarization differentiable.

For cropped text recognition, CRNN~\cite{Shi2017_CRNN} combined CNN, BiLSTM, and CTC. RARE~\cite{Shi2017_RARE} and ASTER~\cite{Shi2019_ASTER} jointly trained a thin-plate-spline rectifier and an attention-based recognizer using transcription supervision. ABINet~\cite{Fang2021_ABINet} added iterative language modeling; TrOCR~\cite{Li2023_TrOCR} used pretrained vision and language Transformers; and SVTR~\cite{Du2022_SVTR} replaced the CNN--RNN backbone with a visual Transformer. Training data scale and curation also substantially affect recognition~\cite{Baek2019_WhatIsWrong,Jiang2023_MAERec}.

Separately trained detectors and recognizers can duplicate feature extraction, while localization errors affect the image regions available for transcription. End-to-end spotters couple the two tasks through shared features and joint supervision. FOTS~\cite{Zhou2018_FOTS} and Mask TextSpotter~\cite{Lyu2018_MaskTextSpotter} combined both tasks in one network; Mask TextSpotter used character segmentation to recognize irregular text. ABCNet~\cite{Liu2020_ABCNet} represented curved text with Bezier curves, while DeepSolo~\cite{Ye2023_DeepSolo} jointly predicted geometry and characters through point queries.

Sequence generation further connected localization and recognition through a common output representation. SPTS~\cite{Peng2022_SPTS} generated a point and its transcription for each instance. To reduce the cost of long autoregressive sequences, SPTS v2~\cite{Liu2023_SPTSv2} generated instance points first, then recognized instances in parallel with shared decoder parameters. UNITS~\cite{Kil2023_UNITS} used prompts to select points, boxes, or polygons as geometric outputs alongside transcriptions. OmniParser~\cite{OmniParser2024} extended sequence modeling to table recognition and downstream key information extraction (\S\ref{sec:vlm_shift}).

Task-specific recognition continues to develop: CPPD and IPAD study efficient decoding~\cite{Du2025_CPPD,Yang2025_IPAD}; SMTR addresses text longer than training instances~\cite{Du2025_SMTR}; LMIM uses linguistically informed self-supervision~\cite{Zhang2025_LMIM}; UnionST improves text synthesis~\cite{Ye2026_UnionST}; and WATERec targets artistic typography~\cite{Ye2026_WATER}. SCUT-EPT highlights long text, editing marks, and noisy backgrounds~\cite{Zhu2019_SCUTEPT}. EasyOCR~\cite{EasyOCR2020}, docTR~\cite{DocTR2021}, MMOCR~\cite{MMOCR2021}, and PaddleOCR~\cite{Du2020_PPOCR} package components into multilingual pipelines, with PaddleOCR also covering layout, tables, and formulas.

Document pretraining addressed relationships beyond isolated text recognition. LayoutLM and its successors~\cite{Xu2020_LayoutLM,Xu2021_LayoutLMv2,Huang2022_LayoutLMv3}, DocFormer~\cite{Appalaraju2021_DocFormer}, and UDOP~\cite{Tang2023_UDOP} combine upstream OCR text, layout coordinates, and visual features into shared representations for document understanding. OCR supplies the transcriptions, while these models learn how text, position, and appearance interact in downstream tasks. Document reconstruction also requires relating recognized content to its layout. The cost and error propagation of the external OCR stage also motivated direct image-to-sequence approaches such as Donut~\cite{Kim2022_Donut} (\S\ref{sec:vlm_shift}).

\subsection{Lessons and Transition to Generative Parsing}\label{sec:pre_vlm_ceiling}

Faithful transcription of a complete page requires accurate local recognition and recovery of the visible structural relations among regions. Errors can therefore arise within individual components and when their outputs are assembled. Modular pipelines offer independently testable and upgradable components, supporting reliability and cost control. Their coordination challenges depend on representation sharing, joint training, and feedback~\cite{Du2020_PPOCR,Long2021_SceneText}.
(1)~\emph{Error propagation:} missed detections or incorrect layout labels can affect later recognition and assembly, while global correction across components is difficult~\cite{Kim2022_Donut,Wei2024_GOT}.
(2)~\emph{Training fragmentation:} different annotations and objectives complicate page-level optimization; shared encoders and multitask training help reuse data and representations~\cite{Tang2023_UDOP}.
(3)~\emph{Output integration:} coherent documents require aligning regions, text, reading order, hierarchy, and specialized markup through interfaces and postprocessing~\cite{Blecher2023_Nougat,Wei2024_GOT}.

The Deep Learning Era brought both learned representations and closer integration of OCR tasks. Joint text spotters reduce detection--recognition handoffs~\cite{Zhou2018_FOTS,Ye2023_DeepSolo}, while sequence prediction connects locations and transcriptions in a common output. The Foundation Model Era builds on these developments through large-scale pretraining and shared generation across more content types and document structures. Whole-page generation and layout-guided regional generation organize this integration differently (Section~\ref{sec:vlm}).

\section{OCR in the Foundation Model Era}\label{sec:vlm}

\begin{figure*}[!t]
\centering
\includegraphics[width=0.95\textwidth]{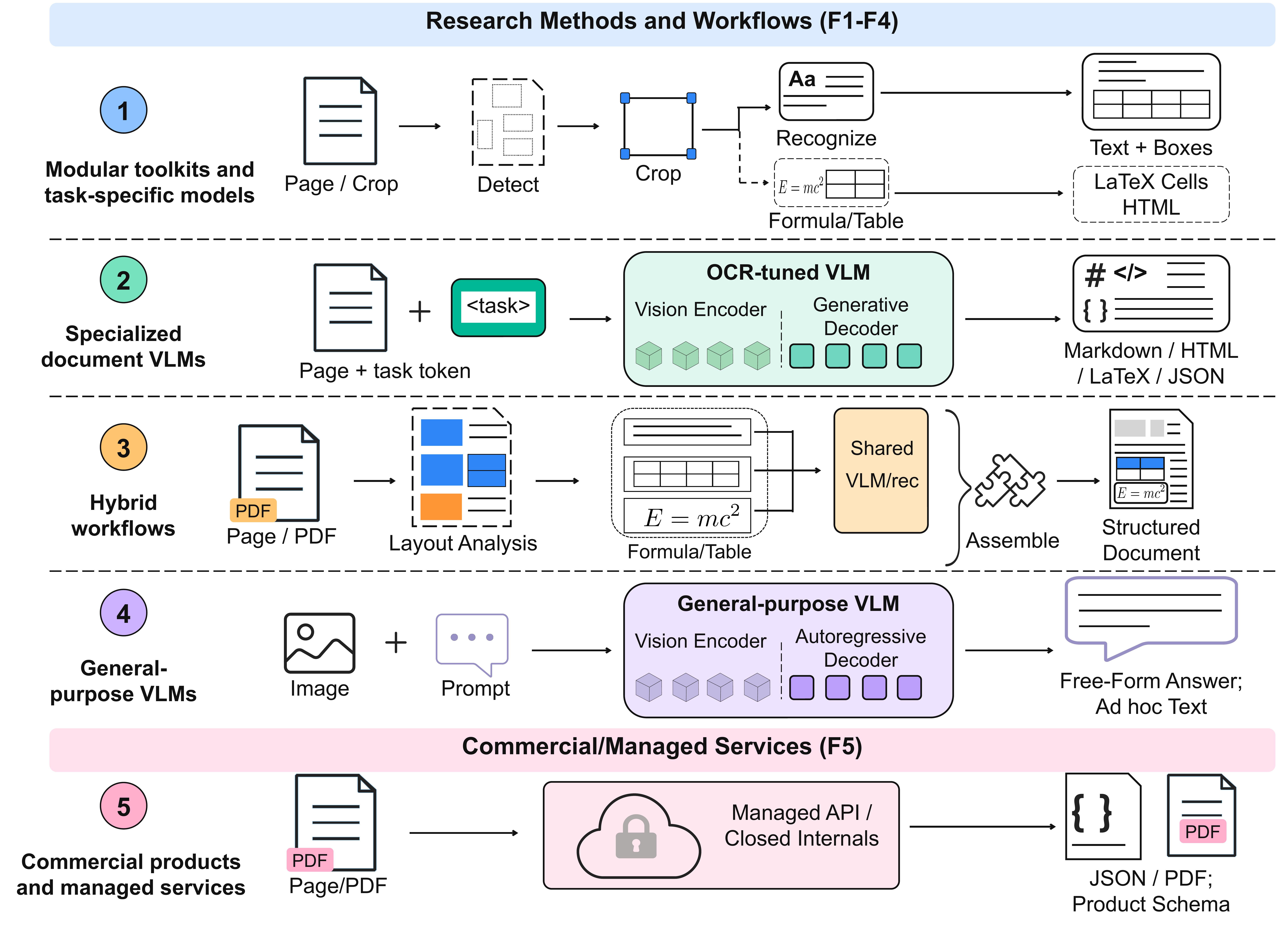}
\caption{Five system families in the OCR ecosystem of the Foundation Model Era, showing how each converts visual input into structured output. Each flow represents a family: \textbf{modular toolkits and task-specific models} combine task-specific prediction with modular processing, \textbf{specialized document VLMs} perform full-page generation with a model trained for documents, \textbf{hybrid workflows} scaffold generation with explicit layout analysis and VLM recognition of individual regions, \textbf{general-purpose VLMs} expose transcription through open prompting, and \textbf{commercial products and managed services} expose closed stacks through product interfaces or managed endpoints. Table~\ref{tab:model_inventory} lists representative systems.}
\label{fig:five_families}
\end{figure*}

In the Foundation Model Era, large-scale pretraining supplies visual and language representations that can be reused across OCR tasks. Building on earlier joint prediction and sequence modeling, generative document models recover text, notation, and structure through shared decoders. We examine three aspects of this unification: which tasks share a model, how their outputs represent content and structure, and how inference coordinates the page and its regions. These models can generate whole pages or operate in stages, as in layout prediction followed by regional recognition. We assess these choices through transcription fidelity, spatial and structural recovery, and inference cost, accounting for differences in architecture, training data, and decoding strategy.

This section traces how generative document parsing emerged (\S\ref{sec:vlm_shift}), organizes the current ecosystem into five coexisting system families and analyzes their architectural differences (\S\ref{sec:vlm_taxonomy}), examines the data and training pipelines that further differentiate them (\S\ref{sec:vlm_design}), and characterizes the limitations that define the current frontier (\S\ref{sec:vlm_limits}). The Generative Parsing track of Fig.~\ref{fig:dl_timeline} charts representative systems from Donut to recent models including OCRVerse~\cite{OCRVerse2026}, Unlimited-OCR~\cite{UnlimitedOCR2026}, and TeleOCR~\cite{NaviDCOCR2026}.

\subsection{Emergence of Generative Document Parsing}
\label{sec:vlm_shift}

Generative document parsing drew on three advances outside OCR. Large-scale pretraining produced transferable representations from unlabeled text~\cite{Devlin2019_BERT} and natural-language supervision of images~\cite{Radford2021_CLIP}; Vision Transformers~\cite{Dosovitskiy2021_ViT} extended Transformer-based pretraining to image patches; and language-model pretraining at scale supplied general-purpose text decoders~\cite{Touvron2023_LLaMA}, while instruction tuning with human feedback improved their ability to follow user instructions~\cite{Ouyang2022_InstructGPT}. These advances expanded the capacity and task coverage of generative OCR.

Three developments adapted general-purpose pretrained models to the demands of document OCR.
First, vision encoders reached the resolution that dense document text demands: Swin Transformer~\cite{Liu2021_SwinTransformer} keeps attention cost tractable at page scale through hierarchical windowing, NaViT~\cite{Dehghani2024_NaViT} processes images at native resolution and aspect ratio, and tiling schemes split a page into subimages so that encoders with a fixed input resolution, such as SigLIP~\cite{Zhai2023_SigLIP}, can be reused. Earlier encoders designed for scene understanding operated at 224--384 pixels~\cite{Dosovitskiy2021_ViT, Radford2021_CLIP} and could miss small characters or fine table lines.
Monkey~\cite{Li2024_Monkey} reused a pretrained encoder on local crops with separate low-rank adapters and a shared resampler, retaining a global image view. TextMonkey~\cite{Liu2024_TextMonkey} introduced shifted-window attention for communication across windows and resampling guided by similarity to reduce redundant visual tokens.
Second, the vision--language model (VLM) architecture emerged as a general framework for multimodal tasks~\cite{Liu2023_LLaVA, Bai2023_QwenVL}: a vision encoder converts the image into visual tokens, a learned projector aligns them with the embedding space of a pretrained large language model, and the LLM generates text conditioned on them. We use \emph{VLM} broadly for models that connect visual representations with language generation. The models discussed here include compact encoder--decoder precursors, such as Donut and Nougat, and recent systems built on pretrained LLM decoders. This architecture gave OCR a visual frontend with high capacity and a unified interface for structured generation.
Third, training data for \emph{whole pages} with structural annotations became available at scale. Rendering markup into images and automatically annotating real documents were established techniques~\cite{Deng2017_Im2Markup, Zhong2020_PubTabNet, Zhong2019_PubLayNet}, but they had supplied only markup for cropped regions or layout boxes for individual modules. Whole pages can be paired with transcription or markup sequences through synthetic document generation and alignment of existing PDFs with structured sources~\cite{Yim2021_SynthDoG, Blecher2023_Nougat}, or through existing OCR pipelines with quality filtering~\cite{Wei2024_GOT, olmOCR2_2025}, supporting \mbox{training on whole pages at scale.}

Early models that map images to sequences without a separate OCR engine, such as Donut~\cite{Kim2022_Donut} and Pix2Struct~\cite{Lee2023_Pix2Struct}, provided important technical foundations. Donut used an encoder--decoder architecture with a decoder initialized from multilingual BART for task-specific document understanding. Pix2Struct used HTML generation from screenshots for pretraining across visual and linguistic tasks. Nougat~\cite{Blecher2023_Nougat} built on Donut's architecture to transcribe complete pages of scientific documents. Its target sequence represents both recognized content and document structure.

OmniParser~\cite{OmniParser2024}, introduced in \S\ref{sec:pre_vlm_dl}, unified prediction across document tasks through explicit spatial representations without a pretrained LLM. It used structure points to organize text spotting, key information extraction, and table recognition, followed by separate contour and text decoders. OmniParser V2~\cite{Yu2026_OmniParserV2} added layout analysis and introduced a shared decoder with separate feedforward branches for structure, localization, and recognition. In its architectural comparison with the same Swin-B backbone, V2 uses 110M parameters versus 144M for OmniParser, with end-to-end text spotting scores of 84.3 versus 84.0 on Total-Text without a lexicon~\cite{Yu2026_OmniParserV2}. This illustrates how partial decoder sharing can reduce model size while preserving text spotting performance. It also used its structured-points-of-thought representation to fine-tune multimodal language models.

In recent OCR VLMs, task unification also draws on vision--language pretraining and multitask training. HunyuanOCR-1.5, for example, combines transcription and document parsing with downstream question answering and translation in one model~\cite{HunyuanOCR15_2026}.

For complete document reconstruction, whole-page generation places content and structure in a common decoding context. Dense pages, however, may require high input resolution and long output sequences. Layout-guided regional generation~\cite{Niu2025_MinerU25,MonkeyOCR2025,Dolphin2025} addresses these costs by identifying regions and decoding their contents separately, retaining an explicit document assembly stage. These two routes draw on shared generative models and training methods and continue to develop in parallel.


\begin{table*}[!t]
\centering
\setlength{\tabcolsep}{2.6pt}
\renewcommand{\arraystretch}{1.0}
\fontsize{8.0}{8.8}\selectfont
\caption{Representative systems in the OCR ecosystem across the five families (\textsf{F1}--\textsf{F5}). Date gives the first documented public appearance of each work, version, or product family (YYYY.MM; YYYY.-- indicates an unavailable month). Outputs list representative content types or formats: Text includes answers and captions for downstream tasks, and Geometry covers boxes, polygons, and points. Params gives official nominal counts for the named model or family (the core VLM in \textsf{F3}); ``--'' marks undisclosed or non-unique values. $\dagger$ marks compact encoder--decoder precursors: Donut for document understanding and Nougat for transcription. Access icons link to official code, model hubs, papers, or services. Full inventory at \url{https://github.com/zhihengli-casia/Awesome-OCR-in-the-Foundation-Model-Era}.}
\label{tab:model_inventory}
\providecommand{\ico}[2]{%
  \IfStrEq{#2}{gh}{\href{#1}{\scalebox{.85}{\faIcon{github}}}}{%
  \IfStrEq{#2}{hf}{\href{#1}{\scalebox{.85}{\faIcon{cube}}}}{%
  \IfStrEq{#2}{web}{\href{#1}{\scalebox{.85}{\faIcon{globe}}}}{%
  \IfStrEq{#2}{doc}{\href{#1}{\scalebox{.85}{\faIcon{book-open}}}}{\href{#1}{\scalebox{.85}{\faIcon{external-link-alt}}}}}}}%
}
\begin{tabular}{@{}>{\centering\arraybackslash}p{5.00cm}
                >{\centering\arraybackslash}p{1.45cm}
                >{\centering\arraybackslash}p{6.00cm}
                >{\centering\arraybackslash}p{2.70cm}
                >{\centering\arraybackslash}p{1.25cm}@{}}
\toprule
\textbf{Model / System} & \textbf{Date} & \textbf{Typical output} & \textbf{Params} & \textbf{Access} \\
\midrule

\rowcolor{gray!8}\multicolumn{5}{@{}c}{\textbf{\textsf{F1}: Modular toolkits and task-specific models}} \\
\rowcolor{gray!4}\multicolumn{5}{@{}c}{\emph{Modular toolkits}} \\
PP-OCRv6~\cite{PPOCRv6_2026} & 2026.06 & Text, Geometry & 1.5--34.5M & \ico{https://github.com/PaddlePaddle/PaddleOCR}{gh} \\
PP-StructureV3~\cite{PaddleOCR2025} & 2025.05 & Markdown, JSON & -- & \ico{https://github.com/PaddlePaddle/PaddleOCR}{gh} \\
Docling~\cite{Docling2024} & 2024.07 & Markdown, JSON & -- & \ico{https://github.com/docling-project/docling}{gh} \\
docTR~\cite{DocTR2021} & 2021.03 & Text, Geometry & -- & \ico{https://github.com/mindee/doctr}{gh} \\
EasyOCR~\cite{EasyOCR2020} & 2020.04 & Text, Geometry & -- & \ico{https://github.com/JaidedAI/EasyOCR}{gh} \\
Tesseract~\cite{Smith2007_Tesseract} & 2005.-- & Text, Geometry & -- & \ico{https://github.com/tesseract-ocr/tesseract}{gh} \\
\rowcolor{gray!4}\multicolumn{5}{@{}c}{\emph{Task-specific models}} \\
SVTRv2~\cite{Du2024_SVTRv2} & 2024.11 & Text & 5.1/11.3/19.8M & \ico{https://github.com/Topdu/OpenOCR}{gh} \\
DocLayout-YOLO~\cite{Zhao2024_DocLayoutYOLO} & 2024.10 & Geometry & 19.5M & \ico{https://github.com/opendatalab/DocLayout-YOLO}{gh} \\
UniMERNet~\cite{UniMERNet2024} & 2024.04 & \LaTeX{} & 107/202/325M & \ico{https://github.com/opendatalab/UniMERNet}{gh} \\
OmniParser~\cite{OmniParser2024} & 2024.03 & Text, Geometry, Markup & -- & \ico{https://github.com/AlibabaResearch/AdvancedLiterateMachinery}{gh} \\
DeepSolo~\cite{Ye2023_DeepSolo} & 2022.11 & Text, Geometry & -- & \ico{https://github.com/ViTAE-Transformer/DeepSolo}{gh} \\
DBNet++~\cite{Liao2022_DBNetpp} & 2022.02 & Geometry & -- & \ico{https://github.com/MhLiao/DB}{gh} \\
TrOCR~\cite{Li2023_TrOCR} & 2021.09 & Text & 62/334/558M & \ico{https://github.com/microsoft/unilm/tree/master/trocr}{gh} \\
ASTER~\cite{Shi2019_ASTER} & 2018.06 & Text & -- & \ico{https://github.com/bgshih/aster}{gh} \\
CRNN~\cite{Shi2017_CRNN} & 2015.07 & Text & 8.3M & \ico{https://github.com/bgshih/crnn}{gh} \\
\midrule

\rowcolor{gray!8}\multicolumn{5}{@{}c}{\textbf{\textsf{F2}: Specialized document VLMs}} \\
HunyuanOCR-1.5~\cite{HunyuanOCR15_2026} & 2026.07 & Text, Markdown, Geometry & 1B & \ico{https://github.com/Tencent-Hunyuan/HunyuanOCR}{gh} \\
OvisOCR2~\cite{OvisOCR2_2026} & 2026.07 & Markdown & 0.8B & \ico{https://huggingface.co/ATH-MaaS/OvisOCR2}{hf} \\
Unlimited-OCR~\cite{UnlimitedOCR2026} & 2026.06 & Text, Geometry & 3B (0.5B act.) & \ico{https://github.com/baidu/Unlimited-OCR}{gh} \\
Qianfan-OCR~\cite{QianfanOCR2026} & 2026.03 & Markdown, HTML, JSON & 4B & \ico{https://huggingface.co/baidu/Qianfan-OCR}{hf} \\
dots.mocr~\cite{DotsMOCR2026} & 2026.03 & Markdown, JSON, SVG & 3B & \ico{https://github.com/rednote-hilab/dots.mocr}{gh} \\
FireRed-OCR~\cite{Wu2026_FireRedOCR} & 2026.02 & Markdown & 2B & \ico{https://github.com/FireRedTeam/FireRed-OCR}{gh} \\
DeepSeek-OCR-2~\cite{DeepSeekOCR2_2026} & 2026.01 & Markdown & 3B (0.5B act.) & \ico{https://github.com/deepseek-ai/DeepSeek-OCR-2}{gh} \\
olmOCR~2~\cite{olmOCR2_2025} & 2025.10 & Text, Markdown & 7B & \ico{https://github.com/allenai/olmocr}{gh} \\
Granite-Docling~\cite{GraniteDocling2025} & 2025.09 & DocTags & 258M & \ico{https://huggingface.co/ibm-granite/granite-docling-258M}{hf} \\
GOT-OCR~2.0~\cite{Wei2024_GOT} & 2024.09 & Text, Markup & 580M & \ico{https://github.com/Ucas-HaoranWei/GOT-OCR2.0}{gh} \\
Nougat$^\dagger$~\cite{Blecher2023_Nougat} & 2023.08 & Markdown & 250/350M & \ico{https://github.com/facebookresearch/nougat}{gh} \\
Donut$^\dagger$~\cite{Kim2022_Donut} & 2021.11 & Text, JSON & 202M & \ico{https://github.com/clovaai/donut}{gh} \\
\midrule

\rowcolor{gray!8}\multicolumn{5}{@{}c}{\textbf{\textsf{F3}: Hybrid workflows}} \\
TeleOCR~\cite{NaviDCOCR2026} & 2026.08 & Markdown, JSON & 1.2B & \ico{https://github.com/caipeng328/TeleOCR}{gh} \\
MonkeyOCRv2-Parsing~\cite{MonkeyOCRv2_2026} & 2026.07 & Markdown, JSON & 0.6/0.7B & \ico{https://github.com/Yuliang-Liu/MonkeyOCRv2}{gh} \\
PaddleOCR-VL-1.6~\cite{PaddleOCRVL16_2026} & 2026.05 & Markdown, JSON & 0.9B & \ico{https://github.com/PaddlePaddle/PaddleOCR}{gh} \\
MinerU~2.5-Pro~\cite{MinerU2025} & 2026.04 & Markdown, JSON & 1.2B & \ico{https://github.com/opendatalab/MinerU}{gh} \\
GLM-OCR~\cite{Duan2026_GLMOCR} & 2026.02 & Markdown, JSON & 0.9B & \ico{https://huggingface.co/zai-org/GLM-OCR}{hf} \\
Youtu-Parsing~\cite{YoutuParsing2026} & 2026.01 & Text, \LaTeX{}, HTML, Geometry & 2.5B & \ico{https://github.com/TencentCloudADP/youtu-parsing}{gh} \\
Dolphin-v2~\cite{Dolphin2025} & 2025.12 & Markdown, JSON & 3B & \ico{https://github.com/bytedance/Dolphin}{gh} \\
MonkeyOCR~\cite{MonkeyOCR2025} & 2025.06 & Markdown, JSON & 3B & \ico{https://github.com/Yuliang-Liu/MonkeyOCR}{gh} \\
\midrule

\rowcolor{gray!8}\multicolumn{5}{@{}c}{\textbf{\textsf{F4}: General-purpose VLMs}} \\
GPT-6~\cite{GPT6_2026} & 2026.10 & Text, Markup & -- & \ico{https://openai.com/index/gpt-6-for-everyone/}{web} \\
Qwen3.8~\cite{Qwen38_2026} & 2026.08 & Text, Markup & 27B & \ico{https://huggingface.co/Qwen/Qwen3.8-27B}{hf} \\
Gemini~3.1~Pro~\cite{Gemini31Pro_2026} & 2026.02 & Text, Markup & -- & \ico{https://deepmind.google/models/gemini/pro/}{web} \\
InternVL3.5~\cite{InternVL35_2025} & 2025.08 & Text, Markup & 1.1--241B & \ico{https://github.com/OpenGVLab/InternVL}{gh} \\
\midrule

\rowcolor{gray!8}\multicolumn{5}{@{}c}{\textbf{\textsf{F5}: Commercial products and managed services}} \\
Mistral OCR~4~\cite{MistralOCR4_2026} & 2026.06 & Markdown, JSON & -- & \ico{https://docs.mistral.ai/studio/document-processing/basic_ocr}{doc} \\
Azure AI Doc Intelligence~\cite{Azure2025_DocumentIntelligence} & 2019.05 & JSON & -- & \ico{https://azure.microsoft.com/en-us/products/ai-services/ai-document-intelligence}{web} \\
Google Document AI~\cite{Google2025_DocumentAI} & 2019.04 & JSON & -- & \ico{https://cloud.google.com/document-ai}{web} \\
Amazon Textract~\cite{AWSTextract2025} & 2018.11 & JSON & -- & \ico{https://aws.amazon.com/textract/}{web} \\
Mathpix~\cite{Mathpix2026_Docs} & 2016.03 & Markdown & -- & \ico{https://mathpix.com/}{web} \\
ABBYY FineReader~\cite{ABBYY2024_FineReader} & 1993.07 & Text, PDF & -- & \ico{https://www.abbyy.com/finereader/}{web} \\
\bottomrule
\end{tabular}
\end{table*}

\subsection{System Architectures and Ecosystem}
\label{sec:vlm_taxonomy}

Recent benchmarks and technical reports have begun categorizing today's coexisting OCR systems into coarse groups such as pipelines, end-to-end systems, and VLMs~\cite{OmniDocBench2025, PureDocBench2026, QianfanOCR2026}. A classical detection--recognition pipeline and a hybrid workflow that pairs a layout detector with generative recognizers are both pipelines, yet differ fundamentally in how they produce output; a specialized document VLM and a general-purpose VLM are both ``end-to-end,'' yet differ in training objective, output consistency, and deployment constraints. We therefore refine the existing groupings into five families by asking two questions of each system.
The first is how recognition and structure are integrated: through task-specific predictions produced within one model or combined by a toolkit or application (Family~1, including joint text spotters and specialized table or formula recognizers); at the region level, where an explicit layout stage selects regions for recognition and subsequent assembly (Family~3); or through structured generation over the whole page (Families~2 and~4).
The second question separates the two families that generate whole pages by training objective: \emph{specialized document VLMs} trained specifically for document OCR form Family~2, while \emph{general-purpose VLMs} with OCR as one capability among many form Family~4.
Commercial products and managed services often disclose too little architectural or training detail for reliable placement on either dimension. We therefore treat their delivery model as a separate category (Family~5).
The families describe OCR workflows; task sharing can occur within each family.
Fig.~\ref{fig:five_families} and Table~\ref{tab:model_inventory} summarize the five families. The table selects historical landmarks, representative current architectures, and widely deployed platforms, and compares their typical outputs, parameter scale, and availability.

\smallskip
\noindent\textbf{Family~1: Modular toolkits and task-specific models.}
This family includes modular toolkits (Tesseract~\cite{Smith2007_Tesseract}, PaddleOCR~\cite{Du2020_PPOCR,PaddleOCR2025}, EasyOCR~\cite{EasyOCR2020}, docTR~\cite{DocTR2021}) that chain detection, recognition, and postprocessing, as well as task-specific models (SVTR~\cite{Du2022_SVTR}, UniMERNet~\cite{UniMERNet2024}, DocLayout-YOLO~\cite{Zhao2024_DocLayoutYOLO}) and joint text spotters. Such models solve defined recognition or localization tasks and can also serve as building blocks in Family~3.
Their task-specific interfaces commonly support constrained or deterministic decoding: detection--recognition pipelines return text with spatial coordinates, table and formula modules emit structured markup, and layout models produce bounding boxes with category labels.
These systems retain task-specific supervision and output formats and are widely used for stable workloads with clear requirements, where they offer predictable outputs and high throughput. Dedicated recognition remains active alongside OCR with foundation models. On WordArt-Bench, WATERec~\cite{Ye2026_WATER} substantially outperforms the VLM baselines evaluated under the same protocol, showing the continuing value of data and architectures tailored to the domain for challenging text distributions.\looseness=-1

\smallskip
\noindent\textbf{Family~2: Specialized document VLMs.}
A specialized document VLM, designed or fine-tuned specifically for document OCR, maps page images to structured output (Markdown, HTML, \LaTeX{}, or JSON), directly targeting the requirement for faithful parsing (Table~\ref{tab:model_inventory}).
These systems commonly combine training focused on documents, visual encoding designed for OCR, and a predefined output schema. Structural relations such as heading hierarchy, table boundaries, and reading order are handled within the generative process, with designs ranging from implicit decoding~\cite{Blecher2023_Nougat} through chain-of-thought layout~\cite{QianfanOCR2026} to grounded multitask parsing~\cite{HunyuanOCR2025}. GOT-OCR shares a language decoder across text and notation formats~\cite{Wei2024_GOT}; HunyuanOCR-1.5 uses task prompts to generate transcriptions, text coordinates, or document markup~\cite{HunyuanOCR15_2026}. Unification therefore concerns both shared model parameters and support for different output requirements.

\smallskip
\noindent\textbf{Family~3: Hybrid workflows.}
These systems bridge Families~1 and~2: an explicit layout stage, run by a dedicated detector (often a Family~1 component) or by the parsing model itself, segments the page into typed regions, and each region is decoded by a VLM recognizer before a postprocessor assembles the results into a structured document~\cite{MinerU2025,MonkeyOCR2025,Dolphin2025}.
Working on bounded regions can shorten generation sequences, localize some errors, and make computational cost easier to budget. Reliability depends on layout recall and crop quality: missed or inaccurate regions may produce omissions that subsequent assembly cannot recover.

Dolphin (2025) generated the page's element sequence with a compact encoder--decoder and then parsed the elements in parallel with element-specific prompts~\cite{DolphinACL2025}. MonkeyOCR (June 2025) built a hybrid workflow on a pretrained LLM decoder: a dedicated detector predicts the block structure, a 3B VLM recognizes each block, and a relation stage orders the blocks~\cite{MonkeyOCR2025}. dots.ocr (July 2025) took the whole-page route, predicting layout and content with one VLM in a single pass~\cite{Li2025_DotsOCR}. Subsequent releases explored smaller recognizers and different degrees of sharing across stages: MinerU2.5 (September 2025) used a 1.2B VLM for layout at reduced resolution and for recognition from crops at native resolution~\cite{Niu2025_MinerU25}; PaddleOCR-VL-1.5 (January 2026) kept the two-stage form with a 0.9B VLM~\cite{PaddleOCRVL2026}; and Dolphin-v2 fine-tuned one VLM for both layout analysis and content parsing, with regional recognition for digital documents and whole-page parsing for photographed ones~\cite{Dolphin2025}. Releases in 2026 kept this structure and moved effort to training and decoding: MinerU2.5-Pro added fine-tuning on difficult samples and PaddleOCR-VL-1.6 progressive post-training~\cite{MinerU2025,PaddleOCRVL16_2026}; MonkeyOCRv2 replaced the reused vision encoder with one pretrained from scratch on document images~\cite{MonkeyOCRv2_2026} (\S\ref{sec:vlm_design}); GLM-OCR and Youtu-Parsing cut sequential decoding steps with multi-token and parallel decoding~\cite{Duan2026_GLMOCR,YoutuParsing2026} (\S\ref{sec:vlm_limits}); and TeleOCR extended the workflow to camera-captured documents~\cite{NaviDCOCR2026}. Recognizers in this family now range from 0.6B to 3B parameters, with most 2026 releases below 1.5B.\looseness=-1

\smallskip
\noindent\textbf{Family~4: General-purpose VLMs.}
General-purpose VLMs (GPT-6~\cite{GPT6_2026}, Gemini~3.1~Pro~\cite{Gemini31Pro_2026}, Qwen3.8~\cite{Qwen38_2026}, InternVL3.5~\cite{InternVL35_2025}) are trained for broad understanding of visual and linguistic content, with OCR as one capability among many.
These models support transcription through prompts, alongside applications such as question answering about images and semantic field extraction. The share of documents in their training mixtures is generally undisclosed. On OmniDocBench v1.6\_full, the general-purpose VLMs on the official leaderboard span 83.76 to 93.70 Overall, from InternVL3.5-241B to Ovis2.6-30B-A3B, and the strongest of them falls within the 93.26--96.91 range of the specialized and hybrid systems in Table~\ref{tab:crossdomain_results}~\cite{OmniDocBenchRepo2026}.

\smallskip
\noindent\textbf{Family~5: Commercial products and managed services.}
Commercial products and managed services (ABBYY, Mathpix, TextIn, Mistral~OCR~4~\cite{MistralOCR4_2026}, Google Document~AI~\cite{Google2025_DocumentAI}, Azure~\cite{Azure2025_DocumentIntelligence}) typically emphasize operational reliability, format control, integration support, and enterprise deployment features. Where documentation describes the processing, it maps onto the preceding families: Document~AI and Azure~AI Document Intelligence expose separate reading, layout, and form or table processors in the manner of Family~1 toolkits, while Mistral~OCR~4 serves one document model that returns Markdown and JSON.\looseness=-1

\subsection{Data and Training Pipelines}
\label{sec:vlm_design}

\begin{figure*}[!t]
\centering
\includegraphics[width=0.95\textwidth]{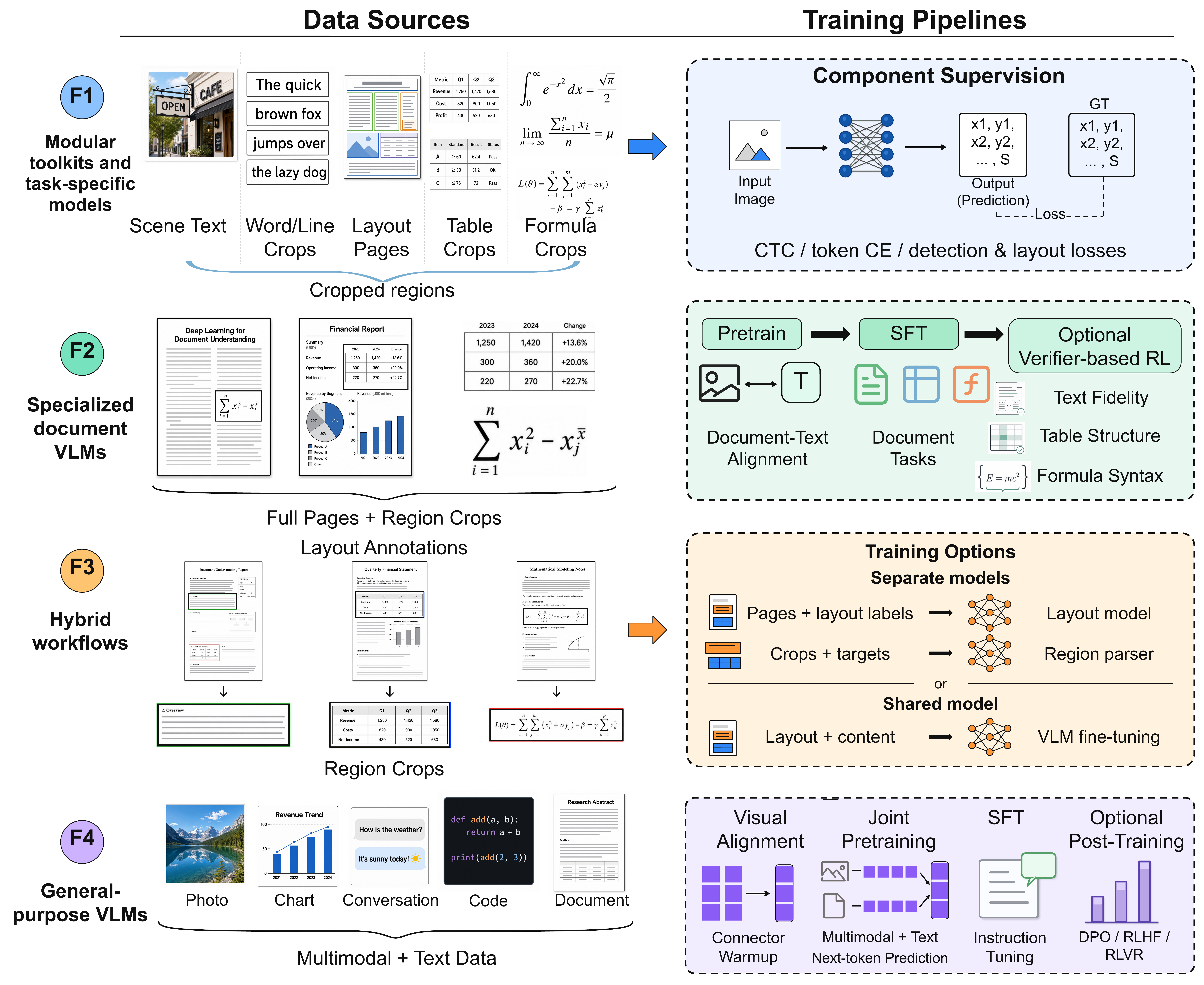}
\caption{Data sources (left) and training pipelines (right) for the first four system families. \textbf{Modular toolkits and task-specific models} use task-specific data and supervision; \textbf{specialized document VLMs} use pages and region crops for pretraining, supervised fine-tuning, and optional reinforcement learning with verifiable rewards; \textbf{hybrid workflows} train separate layout and recognition models or a shared VLM for both tasks; and \textbf{general-purpose VLMs} use broad multimodal corpora with alignment, joint pretraining, instruction tuning, and optional post-training.}
\label{fig:data_training}
\end{figure*}

The capabilities of each system family depend on both its architecture and the data and training signals from which those capabilities are learned. Fig.~\ref{fig:data_training} compares these two dimensions across the four trainable families. Local transcription labels supervise recognition, while page reconstruction also requires reading order, hierarchy, and relations between elements.\looseness=-1

\paragraph{\normalfont\bfseries Data sources}
Data quantity and quality are established performance variables in scene text recognition~\cite{Baek2019_WhatIsWrong, Jiang2023_MAERec}, and the four trainable families draw on progressively broader data (Fig.~\ref{fig:data_training}, left).
Family~1 trains each module on \emph{task-specific data}: scene text images, word and line crops, pages with layout annotations, table crops, and formula crops. Public corpora range from roughly $10^5$ annotated pages or formulas to nearly $10^7$ synthetic word images per task~\cite{Deng2017_Im2Markup,Jaderberg2014_MJSynth,Zhong2019_PubLayNet,Zhong2020_PubTabNet}.
Family~2 assembles \emph{data centered on documents}, with reported training mixtures spanning several million to tens of millions of page and region image--text pairs~\cite{Kim2022_Donut,Blecher2023_Nougat,Wei2024_GOT,DeepSeekOCR2025}. These data combine full pages with region crops through synthetic rendering~\cite{Yim2021_SynthDoG}, engine bootstrapping, deduplication using CLIP~\cite{Radford2021_CLIP}, quality filtering, and curriculum schedules.
Family~3 combines \emph{layout annotations} (pages with layout labels and bounding boxes) with \emph{cropped regions} paired with transcription targets. Representative systems report several million to several tens of millions of mixed training samples, while their subsets for layout supervision range from tens of thousands to a few million pages~\cite{MonkeyOCR2025,Niu2025_MinerU25,PaddleOCRVL2026}.
\begingroup\widowpenalties 2 10000 0\relax
Family~4 trains on \emph{broad multimodal corpora} spanning photographs, charts, conversations, code, and documents, where document understanding is one of many target capabilities. Open technical reports describe hundreds of billions to several trillion pretraining tokens, followed by roughly one to tens of millions of supervised multimodal examples~\cite{Qwen25VL2025,InternVL3_2025,Qwen3VL_2025}; the proportions of documents are generally undisclosed.\par\endgroup

\paragraph{\normalfont\bfseries Training pipelines}
Training methods also diverge across families (Fig.~\ref{fig:data_training}, right).
Family~1 uses \emph{task-specific supervision}: CTC or token cross-entropy for recognizers~\cite{Shi2017_CRNN,Du2022_SVTR}, and detection or layout losses for other modules~\cite{Liao2020_DBNet,Zhong2019_PubLayNet}. Components may be trained separately, while end-to-end spotters share features and jointly optimize localization and recognition~\cite{Lyu2018_MaskTextSpotter,Peng2022_SPTS}.
Family~2 is typically trained in two stages, \emph{pretraining} for alignment between document images and text followed by \emph{supervised fine-tuning} (SFT) on document tasks covering text fidelity, table structure, and formula syntax; a growing number of systems add a third stage of \emph{reinforcement learning with verifiable rewards} (RLVR) that exploits OCR's distinctive property of producing verifiable outputs. These objectives include rewards from executable unit tests for text presence and absence, reading order, tables, formulas, and abnormal repetition in olmOCR2~\cite{olmOCR2_2025} and rewards that account for output format~\cite{FDRL2026}.
Family~3 uses either \emph{separate models} or a \emph{shared VLM} for layout analysis and recognition. With separate models, the layout detector is trained on annotated pages, while region parsers follow Family~2's training pipeline or are reused from existing models. Dolphin-v2 fine-tunes the same VLM for both layout analysis and content parsing~\cite{Dolphin2025}. PaddleOCR-VL-1.5 also uses group relative policy optimization (GRPO) to reduce inconsistency in output style~\cite{PaddleOCRVL2026,Shao2024_GRPO}. With its 1.2B architecture fixed, MinerU2.5-Pro improves table TEDS on OmniDocBench v1.6 from 90.37 to 92.87 by adding supervised fine-tuning on hard examples mixed with earlier training data~\cite{MinerU2025}.
Within Family~3, MonkeyOCRv2 pretrains standalone ViT-Small/Base encoders from scratch on document images, using text and pixel reconstruction decoders that share the visual tokens~\cite{MonkeyOCRv2_2026}. For parsing, it freezes the encoder and connects it to Qwen3-0.6B through an MLP projector. At a long-side input resolution of 448 pixels, adding pixel reconstruction during pretraining raises its S-Parsing pipeline's recognition accuracy on scrambled text from 55.4\% to 72.1\%, showing gains when linguistic context is disrupted~\cite{MonkeyOCRv2_2026}.
Family~4 follows a pipeline with four stages: \emph{visual alignment} (connector warmup), \emph{joint pretraining} on multimodal and text data, \emph{SFT} via instruction tuning, and optional \emph{post-training} with direct preference optimization (DPO), reinforcement learning from human feedback (RLHF), or RLVR.
In downstream document question answering, DocThinker~\cite{Yu2025_DocThinker} uses rule-based reinforcement learning with rewards for answer correctness, supporting-region localization, question reformulation, and output format.

\subsection{Current Limitations}
\label{sec:vlm_limits}

Families~2--4 generate output conditioned on the image. Learned linguistic and layout regularities should support recovery of the content in the source image, but when compression, limited resolution, or unfamiliar layouts weaken the visual evidence, generation may rely on learned priors, producing unsupported content or omitting difficult regions. Localization, reading order, and relations across regions add requirements beyond recognizing the text, and high input resolution with long structured outputs increases computation. Multilingual text and documents with many pages combine these difficulties.

\paragraph{\normalfont\bfseries Hallucination and omission}
Hallucination and omission can arise when visual detail is lost, complex layouts are only partly covered, or decoding errors accumulate in long sequences~\cite{Kim2022_Donut,Wei2024_GOT,OmniDocBench2025}. Published evaluations report fabricated text and phantom table rows, as well as omissions, collapsed columns, and incorrect heading levels~\cite{OmniDocBench2025,PureDocBench2026}. MinerU2.5, MonkeyOCR, PaddleOCR-VL, and Dolphin-v2 use layout analysis and parsing of individual regions to shorten generation sequences and preserve local detail~\cite{Niu2025_MinerU25,MonkeyOCR2025,PaddleOCRVL2026,Dolphin2025}; MinerU2.5-Pro and olmOCR2 strengthen output fidelity through data verification and executable rewards~\cite{MinerU2025,olmOCR2_2025}. The olmOCR2 study reports that a data loader bug excluded blank pages from training and caused hallucinations on blank inputs, an error that benchmark scores did not register, since fixing it and retraining left them unchanged~\cite{olmOCR2_2025}.

\paragraph{\normalfont\bfseries Spatial grounding and structural relations}
Spatial grounding associates each output element, such as text, a formula, a table, or a layout block, with its region in the source image through a bounding box, coordinates, or a region identifier. It makes generated content verifiable against the image and supports reconstruction of reading order, hierarchy, and relations across regions. Its precision is bounded by the visual detail the encoder retains and by coordinate quantization in the output~\cite{HunyuanOCR2025,QianfanOCR2026}. Benchmark evaluations show that general-purpose VLMs remain limited in text localization, layout perception, and parsing of complex elements~\cite{Liu2025_OCRBenchv2}, and plain Markdown, HTML, or \LaTeX{} output generally omits the correspondence between elements and image regions altogether~\cite{Blecher2023_Nougat,Wei2024_GOT}. Hybrid workflows add explicit structure: MonkeyOCR separates structure detection, region recognition, and reading-order prediction through its Where--What--How formulation~\cite{MonkeyOCR2025}; Dolphin-v2 uses absolute coordinates and layout anchors to improve crop precision; and PaddleOCR-VL predicts regions defined by multiple points jointly with reading order and grounded \mbox{text spotting~\cite{Dolphin2025,PaddleOCRVL2026}.}\looseness=1

\paragraph{\normalfont\bfseries Inference cost}
Inference cost arises from visual input at high resolution and from long structured output: dense pages produce many visual tokens, and generating one token at a time introduces latency that grows with output length. GLM-OCR combines multi-token prediction with parallel region recognition~\cite{Duan2026_GLMOCR}, Youtu-Parsing introduces token parallelism and query parallelism over shared page features~\cite{YoutuParsing2026}, and P-MTP progressively increases lookahead depth and dynamically selects speculative lengths during inference~\cite{Xiang2026_PMTP}. Parallel processing of regions and multi-token decoding reduce sequential generation steps, while layout processing, scheduling, verification, and assembly add system overhead.

\paragraph{\normalfont\bfseries Multilingual text}
Language coverage expanded in the earlier eras through dedicated datasets: CASIA-HWDB and IFN/ENIT for Chinese and Arabic handwriting~\cite{Liu2011_CASIAHWDB,Pechwitz2002_IFNENIT}, the ICDAR 2019 MLT challenge for detection and recognition across ten languages in scene images~\cite{Nayef2019_MLT2019}, and XFUND for form understanding in seven languages~\cite{Xu2022_XFUND}. Generative parsers inherit their coverage from the pretrained language model and from the documents added in fine-tuning (\S\ref{sec:vlm_design}), and evaluations show where it falls short: the question-answering benchmark MTVQA reports low accuracy of leading VLMs on text-centric questions in nine languages~\cite{Tang2024_MTVQA}, CC-OCR includes a multilingual reading track~\cite{CCOCR2025}, and MDPBench reports diacritic omissions, character substitutions across writing systems, and errors in ordering text from right to left in document \mbox{parsing~\cite{MDPBench2026}.}\looseness=1

\begin{figure*}[!t]
\centering
\includegraphics[width=\textwidth]{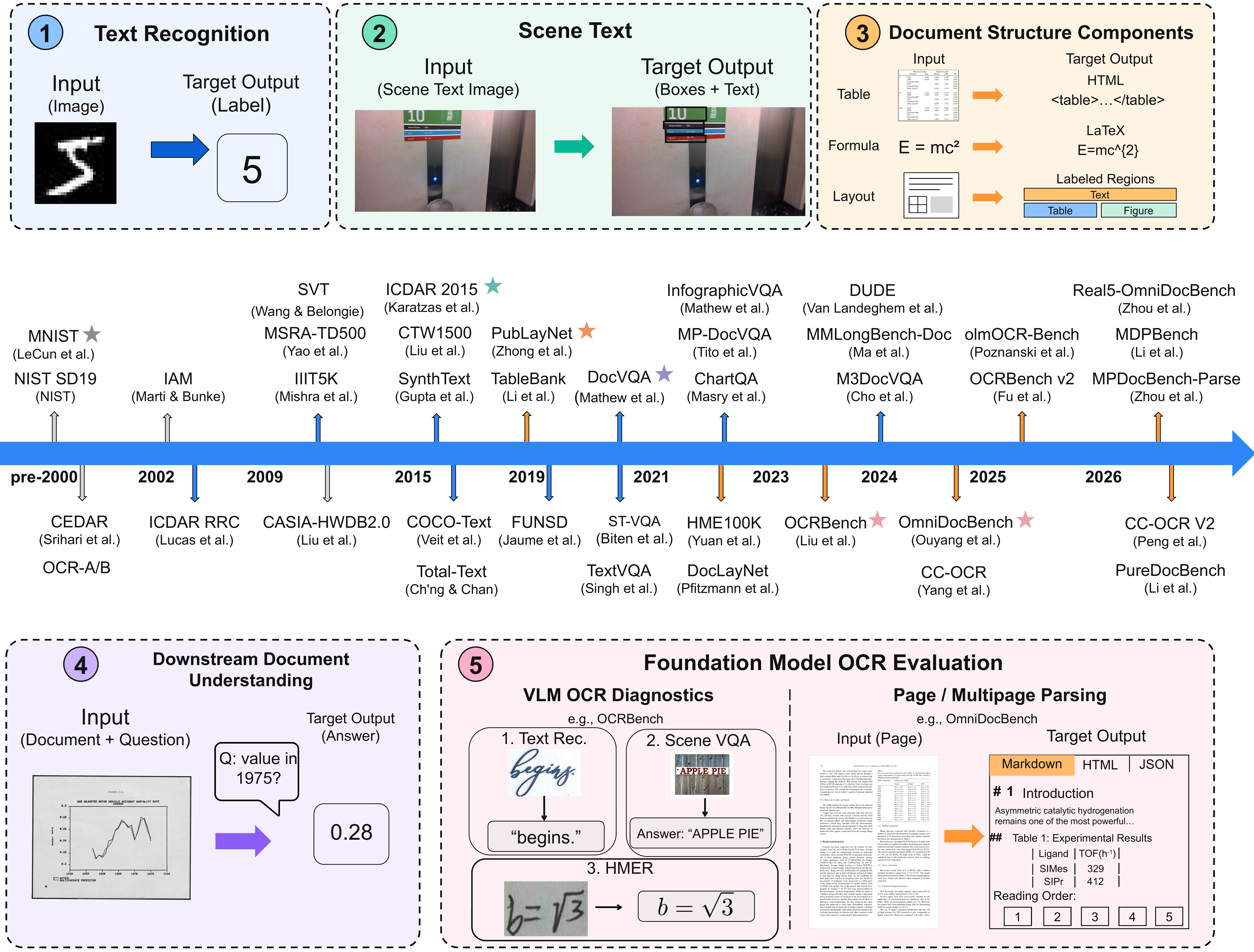}
\caption{Evolution of evaluation relevant to OCR across five stages, combining representative input--output contracts with a chronological timeline of benchmark datasets.
\ding{182}~\emph{Text recognition}: controlled characters, words, and lines (MNIST, IAM, CASIA-HWDB).
\ding{183}~\emph{Scene text}: detection and recognition in natural images (ICDAR, SVT, Total-Text, CTW1500).
\ding{184}~\emph{Document structure components}: tables and formulas as structured markup, plus layouts as labeled spatial regions (PubTabNet, PubLayNet, DocLayNet, HME100K).
\ding{185}~\emph{Downstream document understanding}: question answering about documents (DocVQA, ChartQA, DUDE).
\ding{186}~\emph{Foundation Model OCR Evaluation}, split into two tracks: VLM OCR ability diagnostics (OCRBench and OCRBench~v2) and benchmarks for parsing full pages or multipage documents (OmniDocBench, Real5-OmniDocBench, MDPBench, MPDocBench-Parse, and PureDocBench).
These evaluation directions expand and coexist; VLM diagnostic suites may combine transcription and understanding tasks.
Stars mark selected landmark benchmarks; Table~\ref{tab:benchmark_landscape} lists dataset details and metrics.}
\label{fig:benchmark_evolution}
\end{figure*}

\paragraph{\normalfont\bfseries Long and multipage documents}
Most parsing benchmarks and training data use single pages (Table~\ref{tab:benchmark_landscape}). A document with many pages requires continuing paragraphs and tables across page breaks, recovering a heading hierarchy over the whole document, and maintaining reading order across pages, while visual tokens and output length accumulate with page count. MPDocBench-Parse evaluates these requirements and reports weaknesses in merging content across pages and in recovering hierarchy~\cite{Zhou2026_MPDocBenchParse}. Long-document understanding also addresses the cost of processing many pages: DUDE and MMLongBench-Doc pose questions whose evidence is spread over many pages~\cite{VanLandeghem2023_DUDE,Ma2024_MMLongBenchDoc}, PDF-WuKong parses the document into text and image blocks and then selects evidence by sparse sampling conditioned on the query~\cite{Xie2026_PDFWuKong}, and DocSeeker learns to locate evidence pages through distilled supervision and reinforcement learning~\cite{Yan2026_DocSeeker}.

\section{Evolution of OCR Benchmarks}
\label{sec:experiments}

\setlength{\textfloatsep}{12pt plus 2pt minus 2pt}
\setlength{\dbltextfloatsep}{12pt plus 2pt minus 2pt}

The system evolution of Section~\ref{sec:vlm} changed what OCR benchmarks must measure. The required output has grown from a class label or a character string to a structured reconstruction of a whole page or document, with locations, reading order, tables, formulas, and markup returned together. Evaluation followed this growth in stages, each opened when systems outgrew the measure before it, and none closed by its successors. Section~\ref{sec:exp_benchmarks} traces the five stages; Section~\ref{sec:exp_crossdomain} reads published results as of October 2026 against them.\looseness=-1

\subsection{How OCR Benchmarks Evolved}
\label{sec:exp_benchmarks}

Five stages trace this development (Fig.~\ref{fig:benchmark_evolution} and Table~\ref{tab:benchmark_landscape}): the input expands from isolated characters to complete documents, and the output from class labels through region transcripts and component structures to full-document markup.

The first stage tested \emph{controlled text recognition}. MNIST~\cite{LeCun1998_LeNet} reduced the problem to classifying single digits, while IAM~\cite{Marti2002_IAM} and CASIA-HWDB~\cite{Liu2011_CASIAHWDB} extended evaluation to handwritten characters and complete text lines. These protocols receive the text already located and cropped; finding it defined the next stage.

The second stage brought OCR into natural scenes. The ICDAR Robust Reading competitions~\cite{Karatzas2013_ICDAR,Karatzas2015_ICDAR} and later datasets such as Total-Text~\cite{Chng2017_TotalText} and CTW1500~\cite{Liu2019_CTW1500} established protocols for \emph{scene text detection and end-to-end spotting}, and word-crop benchmarks IIIT5K~\cite{Mishra2012_IIIT5K}, SVT~\cite{Wang2011_SVT}, SVTP~\cite{Phan2013_SVTP}, and CUTE80~\cite{Risnumawan2014_CUTE80} isolated \emph{scene text recognition}. Video text was probed through the ICDAR 2013 video challenge~\cite{Karatzas2013_ICDAR} and BOVText~\cite{Wu2021_BOVText}. The unit of judgment remained the word or region; HierText~\cite{Long2022_HierText} grouped words into lines and paragraphs, while tables, formulas, and page-level structure stayed outside these protocols.\looseness=-1

The third stage, \emph{document structure components}, brought structure into the metric, one component at a time. PubTabNet~\cite{Zhong2020_PubTabNet} evaluated table structure as HTML with TEDS, FinTabNet~\cite{Li2022_FinTabNet} paired cell boxes with logical row--column structure and evaluated cell adjacency, and CROHME~\cite{Mouchere2016_CROHME} and UniMER-Test~\cite{UniMERNet2024} evaluated formula recognition under structural or \LaTeX{} conventions. Layout benchmarks followed a different contract: PubLayNet~\cite{Zhong2019_PubLayNet} and DocLayNet~\cite{Pfitzmann2022_DocLayNet} formulate the task as localization and classification of predefined page regions and report mAP within their respective label spaces. Each component is scored under its own definition of structure, with no shared protocol across components.\looseness=-1

The fourth stage, \emph{document understanding}, judges systems by their answers. DocVQA~\cite{Mathew2021_DocVQA}, ChartQA~\cite{Masry2022_ChartQA}, and DUDE~\cite{VanLandeghem2023_DUDE} pair document images with natural language questions, while FUNSD~\cite{Jaume2019_FUNSD} and CORD~\cite{Park2019_CORD} target key information extraction. These downstream benchmarks evaluate answer or field accuracy and have influenced VLM system design.


\begin{table*}[!p]
\centering
\setlength{\tabcolsep}{2.2pt}
\renewcommand{\arraystretch}{1.00}
\fontsize{6.8}{7.5}\selectfont
\caption{Representative benchmarks across five stages of evaluation relevant to OCR (newest first within each subgroup). Year gives the first public paper or dataset release, or the named version release. Input uses Page for a single page, Multipage for a document, and Mixed for heterogeneous inputs. Geometry denotes boxes or polygons. Main metrics follow official protocols; ED denotes edit distance, CDM evaluates formulas, and TEDS evaluates tables. Scale refers to the named release or split; K/M counts are rounded and test splits are marked. Access icons link to official repositories, dataset pages, or papers. Full list: \href{https://github.com/zhihengli-casia/Awesome-OCR-in-the-Foundation-Model-Era}{project page}.}
\label{tab:benchmark_landscape}
\providecommand{\ico}[2]{%
  \IfStrEq{#2}{gh}{\href{#1}{\scalebox{.85}{\faIcon{github}}}}{%
  \IfStrEq{#2}{hf}{\href{#1}{\scalebox{.85}{\faIcon{cube}}}}{%
  \IfStrEq{#2}{web}{\href{#1}{\scalebox{.85}{\faIcon{globe}}}}{%
  \IfStrEq{#2}{doc}{\href{#1}{\scalebox{.85}{\faIcon{book-open}}}}{\href{#1}{\scalebox{.85}{\faIcon{external-link-alt}}}}}}}%
}
\begin{tabular}{@{}>{\centering\arraybackslash}p{3.45cm}
                >{\centering\arraybackslash}p{1.15cm}
                >{\centering\arraybackslash}p{1.50cm}
                >{\centering\arraybackslash}p{3.60cm}
                >{\centering\arraybackslash}p{3.90cm}
                >{\centering\arraybackslash}p{2.85cm}
                >{\centering\arraybackslash}p{0.80cm}@{}}
\toprule
\textbf{Benchmark} & \textbf{Year} & \textbf{Input} & \textbf{Output} & \textbf{Main metrics} & \textbf{Scale} & \textbf{Access} \\
\midrule

\rowcolor{gray!8}\multicolumn{7}{@{}c}{\textbf{\ding{182} Controlled text recognition}} \\
\rowcolor{gray!4}\multicolumn{7}{@{}c}{\emph{~Character classification}} \\
EMNIST~\cite{Cohen2017_EMNIST}
  & 2017 & Crop & Labels
  & Accuracy & 814K images & \ico{https://www.nist.gov/itl/products-and-services/emnist-dataset}{web} \\
SVHN~\cite{Netzer2011_SVHN}
  & 2011 & Crop & Labels
  & Accuracy & 600K+ images & \ico{http://ufldl.stanford.edu/housenumbers/}{web} \\
MNIST~\cite{LeCun1998_LeNet}
  & 1998 & Crop & Labels
  & Accuracy & 70K images & \ico{http://yann.lecun.com/exdb/mnist/}{web} \\
\rowcolor{gray!4}\multicolumn{7}{@{}c}{\emph{~Handwriting character, line, and word recognition}} \\
CASIA-HWDB 1.x/2.x~\cite{Liu2011_CASIAHWDB}
  & 2011 & Page, Crop & Text, Labels
  & Accuracy, CR/AR & 3.9M chars, 5.1K pages & \ico{https://nlpr.ia.ac.cn/databases/handwriting/Home.html}{web} \\
RIMES 2011 (line)~\cite{Grosicki2011_RIMES}
  & 2011 & Crop & Text
  & CER, WER & 12,111 lines & \ico{https://zenodo.org/records/10812725}{web} \\
IAM~\cite{Marti2002_IAM}
  & 2002 & Crop & Text
  & CER, WER & 13.4K lines & \ico{https://fki.tic.heia-fr.ch/databases/iam-handwriting-database}{web} \\
IFN/ENIT~\cite{Pechwitz2002_IFNENIT}
  & 2002 & Crop & Text
  & Word acc. & 26,459 images & \ico{http://www.ifnenit.com/}{web} \\
\midrule

\rowcolor{gray!8}\multicolumn{7}{@{}c}{\textbf{\ding{183} Scene text}} \\
\rowcolor{gray!4}\multicolumn{7}{@{}c}{\emph{~Detection and spotting}} \\
HierText~\cite{Long2022_HierText}
  & 2022 & Scene & Masks, Hierarchy
  & PQ & 11,639 images & \ico{https://github.com/google-research-datasets/hiertext}{gh} \\
TextOCR~\cite{Singh2021_TextOCR}
  & 2021 & Scene & Text, Geometry
  & Det. F1, Word acc. & 28,134 images & \ico{https://textvqa.org/textocr/}{web} \\
ReCTS~\cite{Zhang2019_ReCTS}
  & 2019 & Scene & Text, Geometry
  & Hmean, 1-NED & 25K images & \ico{https://rrc.cvc.uab.es/?ch=12}{web} \\
LSVT~\cite{Sun2019_LSVT}
  & 2019 & Scene & Text, Geometry
  & Hmean, 1-NED & 450K images (400K weak) & \ico{https://rrc.cvc.uab.es/?ch=16}{web} \\
ArT~\cite{Chng2019_ArT}
  & 2019 & Scene & Text, Geometry
  & Hmean, 1-NED & 10K images & \ico{https://rrc.cvc.uab.es/?ch=14}{web} \\
MLT 2019~\cite{Nayef2019_MLT2019}
  & 2019 & Scene & Text, Geometry, Script
  & Hmean, Script acc., 1-NED & 20K images & \ico{https://rrc.cvc.uab.es/?ch=15}{web} \\
CTW1500~\cite{Liu2019_CTW1500}
  & 2017 & Scene & Geometry
  & P/R/F1 & 1.5K images & \ico{https://github.com/Yuliang-Liu/Curve-Text-Detector}{gh} \\
Total-Text~\cite{Chng2017_TotalText}
  & 2017 & Scene & Geometry
  & P/R/F1 & 1.6K images & \ico{https://github.com/cs-chan/Total-Text-Dataset}{gh} \\
MLT 2017~\cite{Nayef2017_MLT}
  & 2017 & Scene & Geometry, Script
  & Hmean, Script acc. & 18K images & \ico{https://rrc.cvc.uab.es/?ch=8}{web} \\
COCO-Text~\cite{Veit2016_COCOText}
  & 2016 & Scene & Geometry
  & P/R/F1 & 63K images & \ico{https://bgshih.github.io/cocotext/}{web} \\
ICDAR 2015~\cite{Karatzas2015_ICDAR}
  & 2015 & Scene & Geometry
  & P/R/F1 & 1.5K images & \ico{https://rrc.cvc.uab.es/?ch=4}{web} \\
ICDAR 2013~\cite{Karatzas2013_ICDAR}
  & 2013 & Scene & Geometry
  & P/R/F1 & 462 images & \ico{https://rrc.cvc.uab.es/?ch=2}{web} \\
MSRA-TD500~\cite{Yao2012_MSRA}
  & 2012 & Scene & Geometry
  & P/R/F1 & 500 images & \ico{https://tc11.cvc.uab.es/datasets/MSRA-TD500_1}{web} \\
\rowcolor{gray!4}\multicolumn{7}{@{}c}{\emph{~Cropped word recognition}} \\
Union14M-Bench~\cite{Jiang2023_MAERec}
  & 2023 & Crop & Text
  & Word acc., WAICS & 409K images (test) & \ico{https://github.com/Mountchicken/Union14M}{gh} \\
ICDAR 2015 (rec.)~\cite{Karatzas2015_ICDAR}
  & 2015 & Crop & Text
  & Word acc. & 2,077 images (test) & \ico{https://rrc.cvc.uab.es/?ch=4}{web} \\
CUTE80~\cite{Risnumawan2014_CUTE80}
  & 2014 & Crop & Text
  & Word acc. & 288 images (test) & \ico{https://cs-chan.com/downloads_CUTE80_dataset.html}{web} \\
SVTP~\cite{Phan2013_SVTP}
  & 2013 & Crop & Text
  & Word acc. & 639 images (test) & \ico{https://openaccess.thecvf.com/content_iccv_2013/papers/Phan_Recognizing_Text_with_2013_ICCV_paper.pdf}{doc} \\
ICDAR 2013 (rec.)~\cite{Karatzas2013_ICDAR}
  & 2013 & Crop & Text
  & Word acc. & 1,095 images (test) & \ico{https://rrc.cvc.uab.es/?ch=2}{web} \\
IIIT5K~\cite{Mishra2012_IIIT5K}
  & 2012 & Crop & Text
  & Word acc. & 3,000 images (test) & \ico{https://cvit.iiit.ac.in/research/projects/cvit-projects/the-iiit-5k-word-dataset}{web} \\
SVT~\cite{Wang2010_SVT,Wang2011_SVT}
  & 2010 & Crop & Text
  & Word acc. & 647 images (test) & \ico{https://tc11.cvc.uab.es/datasets/SVT_1}{web} \\
\rowcolor{gray!4}\multicolumn{7}{@{}c}{\emph{~Video text spotting}} \\
BOVText~\cite{Wu2021_BOVText}
  & 2021 & Video & Text, Tracks
  & MOTA, MOTP, IDF1 & 2,021 videos, 1.75M frames & \ico{https://github.com/weijiawu/BOVText-Benchmark}{gh} \\
\midrule

\rowcolor{gray!8}\multicolumn{7}{@{}c}{\textbf{\ding{184} Document structure components}} \\
\rowcolor{gray!4}\multicolumn{7}{@{}c}{\emph{~Tables}} \\
PubTables-1M~\cite{Smock2022_PubTables1M}
  & 2021 & Page, Crop & Geometry, Structure
  & AP, GriTS & 947.6K tables & \ico{https://github.com/microsoft/table-transformer}{gh} \\
WTW~\cite{Long2021_WTW}
  & 2021 & Scene, Page & Geometry, Structure
  & F1, TEDS & 14,581 images & \ico{https://github.com/wangwen-whu/WTW-Dataset}{gh} \\
FinTabNet~\cite{Li2022_FinTabNet}
  & 2020 & Crop & Geometry, Structure
  & Adjacency F1 & 112,887 tables & \ico{https://openaccess.thecvf.com/content/WACV2021/html/Zheng_Global_Table_Extractor_GTE_A_Framework_for_Joint_Table_Identification_WACV_2021_paper.html}{doc} \\
PubTabNet~\cite{Zhong2020_PubTabNet}
  & 2019 & Crop & HTML
  & TEDS & 568K tables & \ico{https://github.com/ibm-aur-nlp/PubTabNet}{gh} \\
SciTSR~\cite{Chi2019_SciTSR}
  & 2019 & Crop & Cell graph
  & Structure F1 & 15K tables & \ico{https://github.com/Academic-Hammer/SciTSR}{gh} \\
TableBank~\cite{Li2020_TableBank}
  & 2019 & Page, Crop & HTML, Geometry
  & P/R/F1, BLEU & 417K det., 145K struct. & \ico{https://github.com/doc-analysis/TableBank}{gh} \\
\rowcolor{gray!4}\multicolumn{7}{@{}c}{\emph{~Formulas}} \\
UniMER-Test~\cite{UniMERNet2024}
  & 2024 & Crop & \LaTeX{}
  & ExpRate, ED, CDM & 23,757 images & \ico{https://github.com/opendatalab/UniMERNet}{gh} \\
HME100K~\cite{Yuan2022_HME100K}
  & 2022 & Crop & \LaTeX{}
  & ExpRate & 100K images & \ico{https://ai.100tal.com/dataset}{web} \\
CROHME 2016~\cite{Mouchere2016_CROHME}
  & 2016 & Strokes & Symbols, Relations
  & ExpRate & 1,147 expressions (test) & \ico{https://www.isical.ac.in/~crohme/}{web} \\
Im2LaTeX-100K~\cite{Deng2017_Im2Markup}
  & 2016 & Crop & \LaTeX{}
  & BLEU, Image ED/match & 100K images & \ico{https://github.com/harvardnlp/im2markup}{gh} \\
\rowcolor{gray!4}\multicolumn{7}{@{}c}{\emph{~Layout element detection}} \\
DocLayNet~\cite{Pfitzmann2022_DocLayNet}
  & 2022 & Page & Geometry, Labels
  & COCO mAP & 80,863 pages & \ico{https://github.com/DS4SD/DocLayNet}{gh} \\
DocBank~\cite{Li2020_DocBank}
  & 2020 & Page & Token labels
  & F1 & 500K pages & \ico{https://github.com/doc-analysis/DocBank}{gh} \\
PubLayNet~\cite{Zhong2019_PubLayNet}
  & 2019 & Page & Geometry, Labels
  & COCO mAP & 360K pages & \ico{https://github.com/ibm-aur-nlp/PubLayNet}{gh} \\
\midrule

\rowcolor{gray!8}\multicolumn{7}{@{}c}{\textbf{\ding{185} Document understanding\textsuperscript{\dag}}} \\
\rowcolor{gray!4}\multicolumn{7}{@{}c}{\emph{~Document and scene text VQA}} \\
OCR-Reasoning~\cite{OCRReasoning2025}
  & 2025 & Mixed & Answer, Rationale
  & Accuracy, Reasoning score & 1,069 QA & \ico{https://github.com/SCUT-DLVCLab/OCR-Reasoning}{gh} \\
DUDE~\cite{VanLandeghem2023_DUDE}
  & 2023 & Multipage & Answer, Confidence
  & ANLS, ECE, AURC & 41.5K QA & \ico{https://github.com/duchallenge-team/dude}{gh} \\
ChartQA~\cite{Masry2022_ChartQA}
  & 2022 & Chart & Answer
  & Relaxed acc. & 32.7K QA & \ico{https://github.com/vis-nlp/ChartQA}{gh} \\
DocVQA~\cite{Mathew2021_DocVQA}
  & 2020 & Page & Answer
  & ANLS & 50K QA & \ico{https://www.docvqa.org/}{web} \\
TextVQA~\cite{Singh2019_TextVQA}
  & 2019 & Scene & Answer
  & VQA acc. & 45.3K QA & \ico{https://textvqa.org/}{web} \\
\rowcolor{gray!4}\multicolumn{7}{@{}c}{\emph{~Key information extraction}} \\
XFUND~\cite{Xu2022_XFUND}
  & 2021 & Page & Entities, Relations
  & Entity/Relation F1 & 1.4K forms & \ico{https://github.com/doc-analysis/XFUND}{gh} \\
FUNSD~\cite{Jaume2019_FUNSD}
  & 2019 & Page & Entities, Relations
  & Entity/Relation F1 & 199 forms & \ico{https://guillaumejaume.github.io/FUNSD/}{web} \\
\midrule

\rowcolor{gray!8}\multicolumn{7}{@{}c}{\textbf{\ding{186} Foundation model OCR evaluation}} \\
\rowcolor{gray!4}\multicolumn{7}{@{}c}{\emph{~VLM OCR ability diagnosis}} \\
CC-OCR v2\textsuperscript{\dag}~\cite{CCOCRv2_2026}
  & 2026 & Mixed & Text, Structure, Geometry, Fields, Answer
  & Task scores, Mean & 7,093 samples & \ico{https://github.com/eioss/CC-OCR-V2}{gh} \\
OCRBench v2\textsuperscript{\dag}~\cite{Liu2025_OCRBenchv2}
  & 2024 & Mixed & Task outputs
  & Task scores, Total & 10K QA & \ico{https://github.com/Yuliang-Liu/MultimodalOCR}{gh} \\
CC-OCR\textsuperscript{\dag}~\cite{CCOCR2025}
  & 2024 & Mixed & Text, Structure, Geometry, Fields
  & Task scores, Mean & 7,058 images & \ico{https://huggingface.co/datasets/wulipc/CC-OCR}{hf} \\
MTVQA\textsuperscript{\dag}~\cite{Tang2024_MTVQA}
  & 2024 & Mixed & Answer
  & Accuracy & 28,607 QA & \ico{https://github.com/bytedance/MTVQA}{gh} \\
OCRBench\textsuperscript{\dag}~\cite{Liu2024_OCRBench}
  & 2023 & Mixed & Task outputs
  & Total score & 1K QA & \ico{https://github.com/Yuliang-Liu/MultimodalOCR}{gh} \\
\rowcolor{gray!4}\multicolumn{7}{@{}c}{\emph{~Full-page parsing}} \\
PureDocBench~\cite{PureDocBench2026}
  & 2026 & Page & Markdown
  & Text ED, CDM, TEDS, Order ED & 1,475 pages, 4,425 images & \ico{https://github.com/zhihengli-casia/puredocbench}{gh} \\
ParseBench\textsuperscript{\dag}~\cite{ParseBench2026}
  & 2026 & Page, Multipage & Markdown, HTML, Structured data, Geometry
  & GTRM, ChartDataPointMatch & 2,078 pages & \ico{https://github.com/run-llama/ParseBench}{gh} \\
OmniDocBench v1.6~\cite{OmniDocBench2025,OmniDocBenchRepo2026}
  & 2026 & Page & Markdown
  & Text ED, CDM, TEDS, Order ED & 1,651 pages & \ico{https://github.com/opendatalab/OmniDocBench}{gh} \\
MDPBench~\cite{MDPBench2026}
  & 2026 & Page & Markdown
  & Text ED, CDM, TEDS, Overall & 3,400 images & \ico{https://github.com/Yuliang-Liu/MultimodalOCR}{gh} \\
Real5-OmniDocBench~\cite{Real5_2026}
  & 2026 & Page & Markdown
  & Text ED, CDM, TEDS, Order ED & 1,355 pages, 6,775 images & \ico{https://huggingface.co/datasets/PaddlePaddle/Real5-OmniDocBench}{hf} \\
\bottomrule
\multicolumn{7}{@{}c}{\textsuperscript{\dag}\,Stage~\ding{185} is a transition through downstream document understanding; marked suites involve semantic extraction or reasoning beyond faithful transcription.}
\end{tabular}
\end{table*}

The fifth stage, \emph{OCR evaluation in the Foundation Model Era}, returned to reconstruction and divides into two tracks. OCRBench~\cite{Liu2024_OCRBench} and OCRBench~v2~\cite{Liu2025_OCRBenchv2} diagnose VLM abilities across transcription, localization, and understanding tasks. OmniDocBench~\cite{OmniDocBench2025,OmniDocBenchRepo2026}, Real5-OmniDocBench~\cite{Real5_2026}, PureDocBench~\cite{PureDocBench2026}, MDPBench~\cite{MDPBench2026}, and MPDocBench-Parse~\cite{Zhou2026_MPDocBenchParse} evaluate reconstruction of complete pages and multipage documents across content fidelity, reading order, and document structure, placing multiple system families on shared pages while retaining metrics for individual components. With this stage the unit of evaluation caught up with the unit of system output: the page, and most recently the document.\looseness=-1

\subsection{What the Benchmarks Reveal}
\label{sec:exp_crossdomain}

\subsubsection{Quantitative Performance Trends}

Published results show high scores on several common OCR benchmarks, with small recent gains on the selected scene text benchmarks. Table structure within pages and physical capture conditions still distinguish systems.


\begin{table}[!t]
\centering
\setlength{\tabcolsep}{3.3pt}
\renewcommand{\arraystretch}{1.05}
\small
\caption{Scene text detection Hmean/F1 (\%).}
\label{tab:scene_detection_results}
\begin{tabular*}{\columnwidth}{@{\extracolsep{\fill}}c c c c c@{}}
\toprule
\textbf{Method} & \textbf{Date} & \textbf{IC15} & \textbf{Total} & \textbf{CTW} \\
\midrule
PSENet~\cite{Wang2019_PSENet}           & 2018.06 & 85.7 & 80.9 & 82.2 \\
TextSnake~\cite{Long2018_TextSnake}     & 2018.07 & 82.6 & 78.4 & 75.6 \\
PAN~\cite{Wang2019_PAN}                 & 2019.08 & 82.9 & 85.0 & 83.7 \\
DBNet~\cite{Liao2020_DBNet}             & 2019.11 & 87.3 & 84.7 & 84.9 \\
TextFuseNet~\cite{Ye2020_TextFuseNet}   & 2020.07 & \textbf{92.1} & 87.1 & 86.6 \\
FCENet~\cite{Zhu2021_FCENet}            & 2021.04 & 86.2 & 85.8 & 85.5 \\
TESTR~\cite{Zhang2022_TESTR}            & 2022.04 & 90.0 & 86.9 & 87.1 \\
TPD-STR~\cite{Kim2025_TPDSTR}           & 2025.02 & 89.4 & \textbf{88.7} & \textbf{88.5} \\
\bottomrule
\end{tabular*}
\end{table}


\begin{table}[!t]
\centering
\setlength{\tabcolsep}{1.4pt}
\renewcommand{\arraystretch}{1.05}
\small
\caption{Scene text recognition accuracy (\%) under MJ+ST, following the comparison compiled in SVTRv2~\cite{Du2024_SVTRv2}. IC13/IC15 use the 857/1811-image subsets.}
\label{tab:str_results}
\begin{tabular*}{\columnwidth}{@{\extracolsep{\fill}}c c c cccccc@{}}
\toprule
\textbf{Model} & \textbf{Date} & \textbf{Params} &
\rotatebox{60}{\textbf{IIIT5K}} &
\rotatebox{60}{\textbf{SVT}} &
\rotatebox{60}{\textbf{IC13}} &
\rotatebox{60}{\textbf{IC15}} &
\rotatebox{60}{\textbf{SVTP}} &
\rotatebox{60}{\textbf{CUTE}} \\
\midrule
CRNN~\cite{Shi2017_CRNN}          & 2015.07 &  8.3\,M & 82.9 & 81.6 & 91.1 & 69.4 & 70.0 & 65.5 \\
ASTER~\cite{Shi2019_ASTER}        & 2018.06 & 27.2\,M & 93.3 & 90.0 & 90.8 & 74.7 & 80.2 & 80.9 \\
ABINet~\cite{Fang2021_ABINet}     & 2021.03 & 36.7\,M & 96.2 & 93.5 & \textbf{97.4} & 86.0 & 89.3 & 89.2 \\
SVTR-B~\cite{Du2022_SVTR}         & 2022.05 & 24.6\,M & 96.0 & 91.5 & 97.1 & 85.2 & 89.9 & 91.7 \\
PARSeq~\cite{Bautista2022_PARSeq} & 2022.07 & 23.8\,M & 97.0 & 93.6 & 97.0 & 86.5 & 88.9 & 92.2 \\
SVTRv2-B~\cite{Du2024_SVTRv2}     & 2024.11 & 19.8\,M & \textbf{97.7} & \textbf{94.0} & 97.3 & \textbf{88.1} & \textbf{91.2} & \textbf{95.8} \\
\bottomrule
\end{tabular*}
\end{table}


\begin{table}[!t]
\centering
\setlength{\tabcolsep}{1.5pt}
\renewcommand{\arraystretch}{1.03}
\scriptsize
\caption{End-to-end scene text spotting F1 (\%). Dates refer to the first public papers. ``None'' is lexicon-free; ``Full'' uses all words in the test set.}
\label{tab:scene_spotting_results}
\begin{tabular*}{\columnwidth}{@{\extracolsep{\fill}}l c c cc cc@{}}
\toprule
& & \textbf{IC15} & \multicolumn{2}{c}{\textbf{Total-Text}} & \multicolumn{2}{c}{\textbf{CTW1500}} \\
\cmidrule(lr){3-3}\cmidrule(lr){4-5}\cmidrule(lr){6-7}
\textbf{Method} & \textbf{Date} & \textbf{None} & \textbf{None} & \textbf{Full} & \textbf{None} & \textbf{Full} \\
\midrule
FOTS~\cite{Zhou2018_FOTS}                         & 2018 & --   & --   & --   & 21.1 & 39.7 \\
Mask TextSpotter v2~\cite{Liao2021_MaskTextSpotterV2} & 2019 & -- & 65.3 & 77.4 & -- & -- \\
TextDragon~\cite{Feng2019_TextDragon}            & 2019 & --   & 48.8 & 74.8 & 39.7 & 72.4 \\
ABCNet~\cite{Liu2020_ABCNet}                     & 2020 & --   & 64.2 & 75.7 & 45.2 & 74.1 \\
ABCNet~v2~\cite{Liu2021_ABCNetv2}                & 2021 & --   & 70.4 & 78.1 & 57.5 & 77.2 \\
TESTR~\cite{Zhang2022_TESTR}                     & 2022 & 65.3 & 73.3 & 83.9 & 56.0 & 81.5 \\
DeepSolo~\cite{Ye2023_DeepSolo}                  & 2022 & 73.8 & 83.6 & 89.6 & 64.2 & 81.4 \\
ESTextSpotter~\cite{Huang2023_ESTextSpotter}     & 2023 & --   & 80.8 & 87.1 & 64.9 & 83.9 \\
DG-Bridge~\cite{Huang2024_BridgeSpotter}         & 2024 & --   & 83.3 & 88.3 & \textbf{69.8} & 83.9 \\
TiCLS~\cite{Jang2026_TiCLS}                      & 2026 & \textbf{79.1} & \textbf{84.6} & \textbf{90.4} & 66.2 & \textbf{88.1} \\
\bottomrule
\end{tabular*}
\end{table}


\begin{table}[!t]
\centering
\setlength{\tabcolsep}{0.65pt}
\renewcommand{\arraystretch}{1.02}
\scriptsize
\caption{Selected OmniDocBench~v1.6\_full scores~\cite{OmniDocBenchRepo2026} as of October 2026. Overall/Text/Table/Fml. are \% ($\uparrow$); Text is $(1-\mathrm{Edit})\!\times\!100$, Table is TEDS, Fml. is CDM, and RO is reading order Edit ($\downarrow$).}
\label{tab:crossdomain_results}
\begin{tabular*}{\columnwidth}{@{\extracolsep{\fill}}l c c c c c c c@{}}
\toprule
\textbf{System} & \textbf{Date} & \textbf{Params} &
\textbf{Overall} & \textbf{Text} & \textbf{Table} & \textbf{Fml.} & \textbf{RO} \\
\midrule

\rowcolor{gray!8}\multicolumn{8}{@{}l}{\emph{Specialized document VLMs}} \\
OvisOCR2~\cite{OvisOCR2_2026} & 2026.07 & 0.8\,B & 96.47 & \textbf{97.35} & 94.58 & 97.49 & \textbf{0.112} \\
Unlimited-OCR~\cite{UnlimitedOCR2026} & 2026.06 & 3\,B & 94.00 & 96.06 & 90.21 & 95.72 & 0.128 \\
Qianfan-OCR~\cite{QianfanOCR2026} & 2026.03 & 4\,B & 93.90 & 96.00 & 90.53 & 95.08 & 0.130 \\
FireRed-OCR~\cite{Wu2026_FireRedOCR} & 2026.02 & 2\,B & 93.26 & 96.30 & 88.04 & 95.44 & 0.131 \\
\midrule
\rowcolor{gray!8}\multicolumn{8}{@{}l}{\emph{Hybrid workflows}} \\
TeleOCR~\cite{NaviDCOCR2026} & 2026.08 & 1.2\,B & \textbf{96.91} & 97.33 & \textbf{96.82} & 96.59 & 0.118 \\
PaddleOCR-VL-1.6~\cite{PaddleOCRVL16_2026} & 2026.05 & 0.9\,B & 96.34 & 96.74 & 94.76 & \textbf{97.53} & 0.128 \\
MinerU2.5-Pro-2605~\cite{MinerUPro2605_2026} & 2026.05 & 1.2\,B & 95.72 & 96.40 & 93.62 & 97.15 & 0.123 \\
GLM-OCR~\cite{Duan2026_GLMOCR} & 2026.02 & 0.9\,B & 95.22 & 95.60 & 92.83 & 97.18 & 0.133 \\
\bottomrule
\end{tabular*}
\end{table}


\begin{table}[!t]
\centering
\setlength{\tabcolsep}{0.7pt}
\renewcommand{\arraystretch}{1.03}
\scriptsize
\caption{Selected Real5-OmniDocBench leaderboard scores~\cite{Real5Repo2026} (\%,$\uparrow$), as of October 2026, using the pinned OmniDocBench~v1.5 protocol. S/W/SP/I/SK = scan/warp/screen/illum./skew; All averages the five condition scores; Worst is their minimum.}
\label{tab:real5_results}
\begin{tabular*}{\columnwidth}{@{\extracolsep{\fill}}l c c c c c c c@{}}
\toprule
\textbf{System} & \textbf{All} & \textbf{S} & \textbf{W} & \textbf{SP} & \textbf{I} & \textbf{SK} & \textbf{Worst} \\
\midrule

\rowcolor{gray!8}\multicolumn{8}{@{}l}{\emph{Modular toolkits and task-specific models}} \\
PP-StructureV3~\cite{PaddleOCR2025} & 64.45 & 84.68 & 59.34 & 66.89 & 73.38 & 37.98 & 37.98 \\
\midrule
\rowcolor{gray!8}\multicolumn{8}{@{}l}{\emph{Specialized document VLMs}} \\
OvisOCR2~\cite{OvisOCR2_2026} & 92.29 & 93.77 & 91.40 & \textbf{93.09} & 92.88 & 90.33 & 90.33 \\
\midrule
\rowcolor{gray!8}\multicolumn{8}{@{}l}{\emph{Hybrid workflows}} \\
PaddleOCR-VL-1.6~\cite{PaddleOCRVL16_2026} & \textbf{93.19} & \textbf{94.74} & 92.48 & 92.78 & \textbf{93.28} & \textbf{92.66} & \textbf{92.48} \\
Xiaomi-OCR-0~\cite{XiaomiOCR0_2026} & 92.73 & 93.73 & \textbf{92.76} & 92.80 & 92.68 & 91.69 & 91.69 \\
TeleOCR~\cite{NaviDCOCR2026} & 90.72 & 92.33 & 90.63 & 89.09 & 90.90 & 90.63 & 89.09 \\
GLM-OCR~\cite{Duan2026_GLMOCR} & 90.32 & 92.67 & 90.68 & 91.75 & 91.12 & 85.39 & 85.39 \\
MonkeyOCRv2-B-Parsing~\cite{MonkeyOCRv2_2026} & 89.22 & 89.49 & 89.70 & 88.40 & 88.54 & 89.97 & 88.40 \\
\midrule
\rowcolor{gray!8}\multicolumn{8}{@{}l}{\emph{General-purpose VLMs}} \\
Doubao-Seed-2.1-Pro~\cite{DoubaoSeed21_2026} & 89.02 & 88.85 & 89.36 & 88.99 & 89.13 & 88.79 & 88.79 \\
\bottomrule
\end{tabular*}
\end{table}

\begingroup\widowpenalty=10000
\textbf{Scene text.}
CTW1500 detection Hmean increased from 75.6 for TextSnake to 88.5 for TPD-STR; under the MJ+ST setting, IIIT5K recognition accuracy increased from 82.9 for CRNN to 97.7 for SVTRv2; and Total-Text F1 without a lexicon increased from 65.3 for Mask TextSpotter v2 to 84.6 for TiCLS (Tables~\ref{tab:scene_detection_results}--\ref{tab:scene_spotting_results}). From the 2022 entries to the latest ones, the gains are 1.4, 0.7, and 1.0 points: most of the distance was covered within the Deep Learning Era, and the open problems moved to harder text distributions and to full pages.\par
\endgroup

\textbf{Document structure components.}
Reported scores are also high on several document structure benchmarks. Reported TEDS scores include 88.3 for EDD on the original PubTabNet test set~\cite{Zhong2020_PubTabNet} and 96.3 for TableMASTER on the ICDAR 2021 test split~\cite{Ye2021_TableMaster}; the different test sets prevent a direct estimate of improvement. On PubLayNet, reported mAP includes 91.0 for Mask R-CNN in the RoDLA evaluation~\cite{Chen2024_RoDLA} and 96.2 for VGT~\cite{Da2023_VGT}, and UniMERNet reached 96.8 CDM on UniMER-Test in a 2025 metric study~\cite{Wang2025_CDM}. These scores characterize individual components; full-page benchmarks additionally assess how their outputs preserve reading order and relations between elements.

\begingroup
\clubpenalty=10000
\textbf{VLM OCR diagnostics.}
Shared diagnostic suites measure the reading ability that general-purpose VLMs expose through prompts. Selected OCRBench results range from 736 for GPT-4o to 931 for Qwen3.5-397B-A17B~\cite{Liu2024_OCRBench,Qwen35_2026}, an increase of 195 points that leaves the leading score 69 points below the nominal ceiling of 1,000. The total aggregates 1,000 question-level judgments spanning text recognition, scene-text and document question answering, key information extraction, and handwritten mathematical expression recognition~\cite{Liu2024_OCRBench}, so a change in the total cannot be attributed to any single ability, and reconstruction of complete pages lies outside the task set.\par
\endgroup

\textbf{Parsing complete pages.}
Benchmarks for parsing complete pages evaluate reconstruction under a common output protocol. On OmniDocBench~v1.6\_full, the selected results span 93.26--96.91 Overall, 95.60--97.35 for text, 88.04--96.82 for tables, and 95.08--97.53 for formulas (Table~\ref{tab:crossdomain_results})~\cite{OmniDocBenchRepo2026,OvisOCR2_2026,NaviDCOCR2026}. Text scores of the eight systems lie within 1.75 points of one another, and table scores spread over 8.78 points, so tables are the component that still separates them. TeleOCR and OvisOCR2 hold the top two positions, one from layout-guided regional generation and one from whole-page generation. Every system above 95 Overall has at most 1.2B parameters, while the 2B to 4B systems score at most 94.00. Within these selected results, the choice between the two designs does not decide the ranking, and larger models do not rank higher.

\textbf{Physical robustness.}
Real5-OmniDocBench evaluates scanning, warping, screen photography, illumination change, and skew over the same 1,355 source pages from OmniDocBench~v1.5. Across the selected systems on its current official leaderboard, overall scores span 64.45--93.19 (Table~\ref{tab:real5_results})~\cite{Real5Repo2026}. The modular toolkit PP-StructureV3 falls from 84.68 on scans to 37.98 under skew, while every listed generative system stays above 85 in all five conditions, and their overall scores lie within 4.2 points of one another. Each image is a re-capture of one source page under a single acquisition condition~\cite{Real5_2026}, so documents that exist only as camera captures lie outside the benchmark's coverage.

\subsubsection{What Current Metrics Miss}
\label{sec:exp_metric_gaps}

Across protocols, interpretation is limited by three recurring issues: differences in output definitions and matching rules, aggregation across heterogeneous components, and incomplete coverage of content, document conditions, and reconstruction errors.

\textbf{Ambiguity in output definitions and protocols.}
Document structure admits multiple reasonable decompositions and serializations. Borderless tables, aligned fields, forms, and layouts with multiple columns can cross category boundaries, while visually equivalent tables may have different HTML or \LaTeX{} representations. Benchmarks also use different region ontologies, annotation granularities, and matching rules, as illustrated by PubLayNet, DocLayNet, and OmniDocBench~\cite{Zhong2019_PubLayNet,Pfitzmann2022_DocLayNet,OmniDocBench2025,OmniDocBenchRepo2026}. A reported score therefore reflects both the system's reconstruction and the protocol's definition of structure.\looseness=-1

\textbf{Composite score attribution.}
OmniDocBench~v1.6 illustrates the attribution problem: its \emph{Overall} score combines text, table, and formula metrics, while reading order is reported separately~\cite{OmniDocBenchRepo2026}. Similar overall scores can therefore represent different component profiles.

\textbf{Coverage boundaries.}
\begingroup\widowpenalties 2 10000 0\relax
Current benchmarks for complete pages concentrate on text, mathematical formulas, tables, layout regions, and reading order. Charts, flowcharts, chemical structures, and sheet music also encode content and structure that can be recovered from visual evidence; GOT, Logics-Parsing, and OCRVerse already output some of them from page images~\cite{Wei2024_GOT,LogicsParsing2025,OCRVerse2026}. ParseBench extends evaluation to charts by checking sampled data points and their labels~\cite{ParseBench2026}, leaving the completeness of chart reconstruction outside this metric.\par\endgroup

Coverage also varies by language, document length, and capture condition. MDPBench evaluates digital and photographed documents in 17 languages and reports larger errors on non-Latin scripts and photographed pages, particularly among open-source models~\cite{MDPBench2026}. Together with the controlled recaptures in Real5-OmniDocBench~\cite{Real5_2026}, these results motivate reporting performance by language and capture condition. MPDocBench-Parse evaluates cross-page text and table merging and heading hierarchies in English and Chinese~\cite{Zhou2026_MPDocBenchParse}. Extending these document-level tests to more languages would help establish whether multilingual reading also preserves structure across page breaks.

Faithfulness also requires targeted error checks. CHAOS-Bench measures whether models preserve words with deliberately altered characters when visual evidence conflicts with language priors~\cite{HunyuanOCR15_2026}; a missed match can reflect omission, correction, or another substitution. ParseBench separately includes checks for missing, unexpected, and repeated content~\cite{ParseBench2026}. Reporting these error types alongside aggregate text scores would show whether a system alters, omits, or invents content.

\section{Discussion and Outlook}
\label{sec:challenges}

This survey defines OCR through faithful transcription, traces its development from specialist to unified models, and examines recognition and structural recovery using published benchmark results. Looking ahead, three expectations stand out.\looseness=-1

First, as models and agents increasingly use OCR outputs, hallucinations, omissions, and structural errors can affect downstream results. Future work should strengthen verification of content accuracy, completeness, and structural relations against the source image. Second, requirements for accuracy, cost, and control differ across uses, so specialist, unified, and hybrid approaches will continue to coexist. Third, the object of reconstruction will keep widening, from text to the structure of whole documents, from single pages to many, from a few scripts to all writing systems, and from clean scans to whatever a camera captures, and evaluation must widen with it.

OCR's lasting value lies in turning visually recorded information into accurate digital content that can be searched, edited, processed, and reused.

\FloatBarrier
\bibliographystyle{IEEEtran}
\bibliography{references}

\end{document}